\documentclass[preprint,11pt]{elsarticle}

\usepackage{graphicx}
\usepackage{amssymb}
\usepackage{lineno}
\newcommand{\figref}[1]{Figure~\ref{fig:#1}} 
\newcommand{\tblref}[1]{Table~\ref{tbl:#1}}

\newcommand{\eqref}[1]{Equation~\ref{eq:#1}}
\usepackage{setspace}
\usepackage{multirow}
\usepackage{lineno,hyperref}
\usepackage{tabularx}
\usepackage{subcaption}

\journal{Journal Name}

\begin{document}

\begin{frontmatter}



\title{Pervasive Lying Posture Tracking}
\author{Paratoo Alinia\textsuperscript{1,2}, Ali Samadani\textsuperscript{1}, Mladen Milosevic\textsuperscript{1}, Hassan Ghasemzadeh\textsuperscript{2}, and Saman Parvaneh\textsuperscript{1}}

\address{\textsuperscript{1} Philips Research North America, Cambridge, Massachusetts, USA}

\address{\textsuperscript{2} School of Electrical Engineering and Computer Science, Washington State University, Pullman, Washington, USA}

\begin{abstract}
Automated lying-posture tracking is important in preventing bed-related disorders such as pressure injuries, sleep apnea, and lower-back pain. Prior research studied in-bed lying posture tracking using sensors of different modalities (e.g., accelerometer and pressure sensors). However, there remain significant gaps in research about how to design efficient in-bed lying posture tracking systems. These gaps can be articulated through several research questions as follows. First, can we design a single-sensor, pervasive, and inexpensive system that can accurately detect lying postures? Second, what computational models are most effective in the accurate detection of lying postures? Finally, what physical configuration of the sensor system is most effective for lying posture tracking? To answer these important research questions, in this article, we propose a comprehensive approach to design a sensor system that uses a single accelerometer along with machine learning algorithms for in-bed lying posture classification. We design two categories of machine learning algorithms based on deep learning and traditional classification with handcrafted features to detect lying postures. We also investigate what wearing sites are most effective in accurate detection of lying postures. We extensively evaluate the performance of the proposed algorithms on nine different body locations and four human lying postures using two datasets. Our results show that a system with a single accelerometer can be used with either deep learning or traditional classifiers to accurately detect lying postures. The best models in our approach achieve an F-Score that ranges from $95.2$\% to $97.8$\% with $0.03$ to $0.05$ coefficient of variation. The results also identify the thighs and chest as the most salient body sites for lying posture tracking. Our findings in this article suggest that because accelerometers are ubiquitous and inexpensive sensors, they can be a viable source of information for pervasive monitoring of in-bed postures.

\end{abstract}


\end{frontmatter}


\section{Introduction}
	Keeping track of in-bed lying postures and the transitions between them provide useful clinical information about the patients' mobility \cite{azuh2016benefits, hoyer2016promoting}, risk of developing hospital-acquired pressure injuries \cite{neilson2014using}, hidden death in epilepsy \cite{kloster1999sudden}, and quality of sleep \cite{lee2009sleep}. In-bed posture and activities of the patients in hospitals are usually monitored manually and through visual observation, which is a labor- and cost-intensive task. Therefore, researchers have proposed to monitor in-bed postures continuously and non-obtrusively using wearable sensors.
	
	Automatic in-bed lying posture tracking systems have been developed using data collected from sensors of different modalities, such as accelerometers \cite{wrzus2012new,zhang2015monitoring,kwasnicki2018lightweight}, load cell sensors \cite{austin2012unobtrusive}, pressure sensors \cite{pouyan2013continuous,yousefi2011bed}, infrared cameras \cite{cary2016examining}, electrocardiogram waveforms \cite{lee2013estimation}, and multi-modal systems \cite{chang2011wireless,wai2010situation,huang2010multimodal}. Pressure mats and load sensor systems impose a high cost to the end-users and often require calibration. The camera-based systems usually encounter setup and privacy issues from the end-users and are more difficult to analyze than the wearable sensors \cite{lee2015sleep}. Other works have utilized multiple wearable sensors on different body locations for continuous lying-posture detection, which imposes discomfort to the end-users and impede long-term monitoring. To address these issues, we develop a traditional machine learning (ML) model for lying posture detection using a single accelerometer sensor, which is is an ensemble of decision tree classifiers with hand-engineered time-domain features.
	
	Deep learning (DL) has emerged as the leading approach in the field of computer vision, voice recognition, and natural language processing in recent years. Deep neural networks are known as learners of high-level features for a specific problem domain. This makes DL models suitable models for human posture estimation. Moreover, DL models tend to reduce the overhead of feature engineering compared to traditional machine learning models \cite{goodfellow2018atrial}. To date, no studies have explored the possibility of utilizing deep neural networks to acceleration-based lying posture tracking. We develop a deep learning model for lying posture detection using a single accelerometer sensor to investigate the possibility of replacing traditional feature-based machine learning models with deep neural networks, therefore, reduce the burden of feature-engineering. Our deep learning model, adaptive long short-term memory network (AdaLSTM), is a long short-term memory network (LSTM) that uses an adaptive learning rate method with a decaying learning rate schedule.  
	
	More specific contributions of this paper are as follows. We (1) investigate the efficacy of a single accelerometer for lying-posture tracking using feature engineered machine learning models and deep LSTM networks; (2) identify the set of optimal time-domain features for accurate lying posture detection; (3) compare traditional machine learning with deep learning in recognition of lying postures; (4) evaluate nine different body sites to determine the most appropriate site to attach the accelerometer for accurate lying posture tracking. Our main findings as a result of our extensive experimental evaluation are as follows. We identified amplitude, mean, minimum, and maximum values of the lateral and vertical axes of the accelerometer sensor as the optimal set of time-domain features for accurate lying posture tracking. The proposed ensemble tree classifier achieved $97.1\%$ F1-Score and $0.14$ CoV when applied to the data from the sensor on the right thigh, and AdaLSTM obtained $95.3\%$ F1-Score and $0.15$ CoV when applied to the data from the chest sensor. These results demonstrated the thighs and the chest as the optimum location for the accelerometer sensor for accurate lying posture tracking.

		\section{Background \& Related Studies}
	In this section, we discuss the previous studies on lying posture tracking using wearable accelerometer sensors. We divide these studies based on the number of wearable sensors into 1) multi-sensor and 2) single-sensor lying posture tracking.
	
	\subsection{Multi-Sensor Lying Posture Tracking}
	
	A study in \cite{kwasnicki2018lightweight} by Kwasnicki et al. proposed a lightweight sensing platform for monitoring sleep quality and posture using three wearable accelerometer sensors that were placed on both arms and the chest. They applied a K-nearest neighbor, naive Bayes, and decision tree classifiers on the mean and variance of each axis of the signal from all three accelerometer sensors. Their models achieved $99.5\%$ average accuracy in detecting the four major lying postures (i.e., lying supine, prone, and laterals). In another study fallmann et al. proposed a lying posture detection algorithm using three accelerometer sensors on the chest and the legs. Their algorithm first, classified the postures using the acceleration-moving variance method, into stable and non-stable time windows, then classified the features into the postures prone, supine, and laterals. Their model achieved an average accuracy of $83.6\%$ \cite{fallmann2017wearable}. Moreover, Wrzus et al. \cite{wrzus2012new} developed a $99.7\%$ accurate classification model using chest and thigh accelerometry data based on the angular orientation of the upper body along the vertical axis to classify lying postures. 
	
	The above-mentioned algorithms which rely on data from multiple sensors attached to the different locations on the user's body cause discomfort and limit usability, especially for long-term monitoring during sleep. Moreover, the possibility of sensor rotation during sleep might alter the angular axes of the sensors relative to each other, therefore decrease the accuracy of the orientation-based lying posture tracking. In this paper, we proposed lying posture detection algorithms that only use data from a single accelerometer sensor, which can be placed on one of the nine different body locations, including chest, thighs, ankles, arms, and wrists. The proposed models using a single accelerometer are more comfortable to the end-users, therefore, are favored over those using multi-sensor. 
	
	\subsection{Single-sensor Lying Posture Tracking}
	
	In a study by Razjouyan et al. in \cite{razjouyan2017improving}, authors developed a lying posture detection algorithm based on a single accelerometer sensor on the chest of the user. They used a logistic regression model on 43 time-domain features which were extracted from the magnitude of the tri-axial accelerometer signal. The proposed model achieved $87.8\%$ accuracy in detecting the lying postures supine, prone, and laterals for 21 users. In another study \cite{zhang2015monitoring} by Zhang et al., the authors assessed the possibility of using a single accelerometer sensor on the chest to detect the lying posture during sleep. They used linear discriminant analysis (LDA) classifier on the mean value of each axis of the acceleration signal. They achieved an overall accuracy of $99\%$ for classifying lying postures (lying supine, prone, and laterals). However, the authors of this study did not assess the effect of sensor location on the accuracy of lying posture tracking. In another study, Chang et al. developed a system that captured information about sleep events using a smartwatch. Their system distinguished sleep postures supine, prone, and laterals at an average precision of $96\%$. Their proposed algorithm detected the sleeping postures by combining the position of both hands and classification of features using a template-based Euclidean distance matching approach \cite{chang2018sleepguard}. However, the performance of such a model is highly dependent on the quality of pre-defined hand positions and sleep posture templates. Furthermore, the possible sensor rotations during sleeping might affect the accuracy of hand position recognition; and therefore affect the lying posture detection. Moreover, Jeng et. al \cite{jeng2019wrist} proposed a sleep position detection algorithm that achieved $90\%$ accuracy in detecting postures supine, prone and laterals using the data collected from an accelerometer sensor on the wrist of the users. Their proposed model applied a support vector machine classifier with a linear kernel and a random forest of 100 trees on the mean value of the signal.

	\section{Methodologies}
	
\begin{figure*}[ht]
		\centering
		\includegraphics[width=\linewidth]{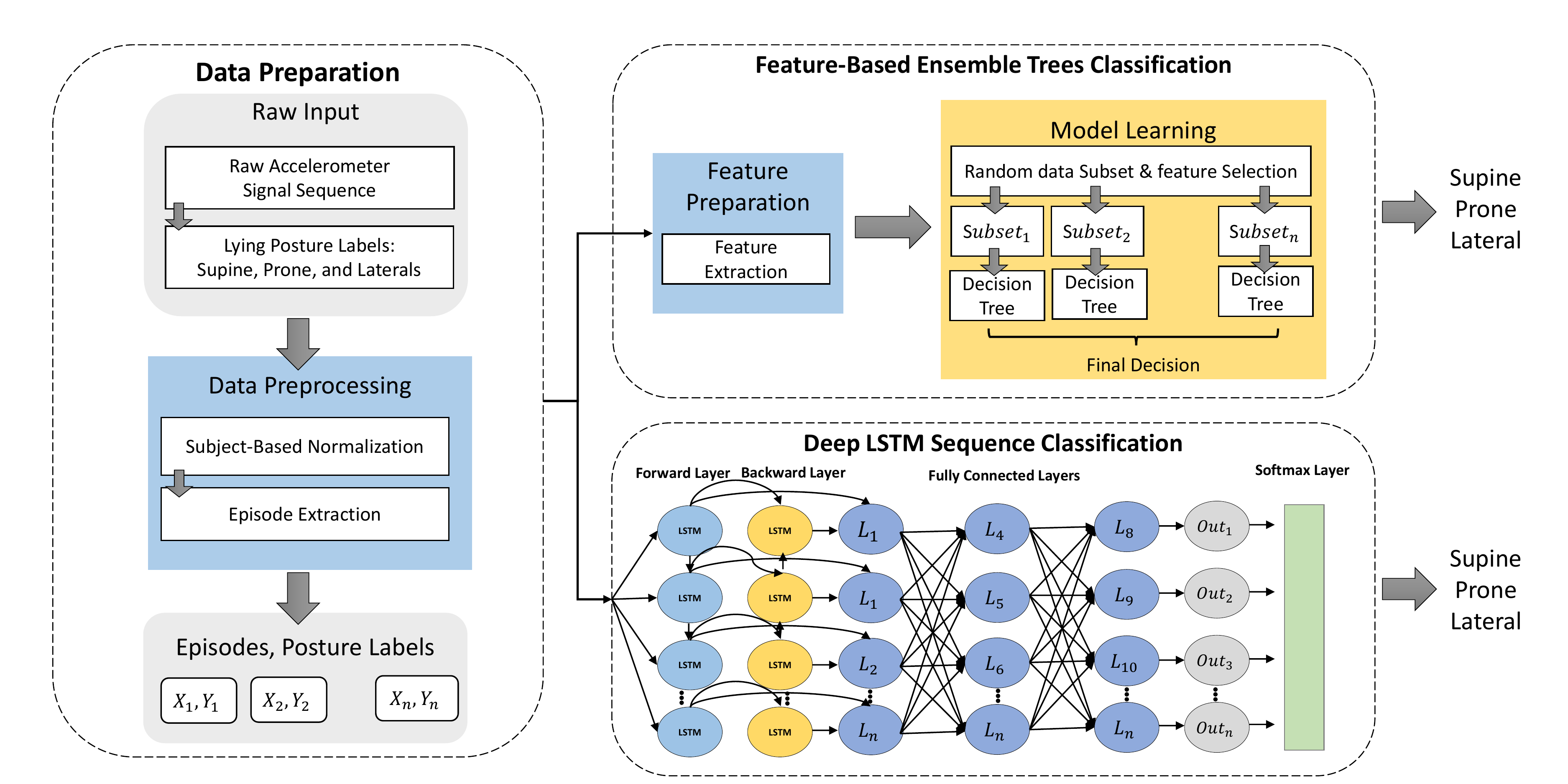}
		\caption{The process of training feature-based ensemble trees and Deep LSTM classifiers.}
		\label{fig:model}
	\end{figure*}

	\figref{model} shows the overall architecture of the proposed ML and DL lying posture tracking. The overall training process includes two steps of data preparation and model development. 
	
	\subsection{Data Preparation}
	We define an episode of data as a sequence of signals collected from one subject while performing a run of a specific posture (e.g., lying supine). The raw accelerometer signal and lying posture labels (e.g., supine, prone, and lateral) are fed into the data processing unit. The processing unit normalizes the signal to remove possible subject-based variations and segments it into different lying posture episodes based on the labels.

	\subsection{Traditional Lying Posture Tracking}
	The proposed traditional lying posture tracking consists of two main steps: (1) feature preparation in which an exhaustive set of features are extracted from each input episode;  and (2) ensemble model learning, which trains an ensemble of 100 decision trees on the features and lying posture labels. 
	
	\subsubsection{Feature Preparation}
	We extract 48 time-domain features from a sliding window over each episode of data (lying supine, prone, and on the left side) with a 50\% overlap. We set the window size to the minimum episode length in the training dataset (e.g.,  96 samples equal to 3.8 seconds). The selected features, shown in \tblref{features}, are proven to be useful in human posture and activity recognition applications such as amplitude, mean, standard deviation, and angle of the signal \cite{mannini2010machine, saeedi2014cost}. We compute a set of meta-features $f_i = \{f^1_i, f^2_i, ..., f^{48}_i\}$ for $i_{th}$ episode by averaging of all the extracted features from multiple windows over that episode. 
	
	\begin{table}[ht]
		\scriptsize
		\caption{Extracted time-domain features. $E(.)$ represents the expected value of the input variable. Functions $min(.), max(.), mean(.), median(.),$ $tan(.)$, $size(.)$ compute the minimum, maximum, average, median, tangent, and size of an input vector.}
		\centering
		\begin{tabular}{|c|l|l|c|}
			\hline
			Feature & Description&Computation for signal $S$ & Number \\ \hline
			AMP     & Peak amplitude & $max(S) - mean(S)$ & 1--3    \\ \hline
			MED     & Median & median(S) & 4--6\\ \hline
			MEAN    & Mean value & $\mu=\frac{\sum_{i=1}^{N}S_i}{N}$                    & 7--9    \\ \hline
			MAX     & Maximum value & $max(S)$            & 10--12  \\ \hline
			MIN     & Minimum value  & $min(S)$          & 13--15  \\ \hline
			VAR     & Variance& $v=\frac{\sum_{i=1}^{N}|s_i-\mu|^2}{N-1}$           & 16--18  \\ \hline
			STD     & Standard deviation  &  $\sigma=\sqrt{\frac{\sum_{i=1}^{N}|S_i-\mu|^2}{N-1}}$  & 19--21  \\ \hline
			RMS    & Root mean square& $\frac{\sum_{i=1}^{N}S_i^2}{N}$& 22--24  \\ \hline
			P2P     & Peak to peak &    $max(S) - min(S)$     & 25--27  \\ \hline
			ZCR     & Zero crossing rate&  $\frac{size(\{S_i|S_i==0, i=1,2,.., N\})}{N}$   &  28--30  \\ \hline
			ENT     & Entropy  &      $-\sum_{i=1}^{N}S_i log(S_i)$        & 31--33  \\ \hline
			SKN     & Skewness & $s=\frac{E(S-\mu)^3}{\sigma^3}$             & 34--36  \\ \hline
			KRT     & kurtosis &   $k=\frac{E(S-\mu)^4}{\sigma^4}$           & 37--39  \\ \hline
			MAG     & Mean Magnitude &   $M=\frac{\sum_{i=1}^{N}\sqrt{S_ix^2+S_iy^2+S_iz^2}}{N}$          & 40     \\ \hline
			ENG     & Energy & $e=\sum_{i=1}^{N}S_i^2$ & 41  \\ \hline
			RNG     & Range &   $r = max(S)-min(S)$          & 42--44  \\ \hline
			ANG     & Angle &   $a = max(tan(\frac{S_z}{S_x^2+S_y^2}))$          & 45     \\ \hline
			MAD     & Mean absolute deviation& $m=\frac{\sum_{i=1}^{N}|s_i-\mu|}{N}$   & 46--48  \\ \hline
		\end{tabular}
		
		\label{tbl:features}
	\end{table}
	
	\subsubsection{Ensemble Model Learning}
	
	A total of 48 features are obtained from the previous step. We train an ensemble of 100 decision trees on the features and lying posture labels. We choose the bagging technique as the ensemble method to reduce the variance of the decision tree and overfitting to the existing data. A decision tree is selected as the weak learner because of the high dimension of the input features. As shown in \figref{model}, we fit 100 decision trees on 100 random subsets of the original dataset with a randomly selected subset of features to minimize the correlation between individual trees. The final prediction is the majority voting on the decision of the individual trees \cite{baskin2017bagging}.
	
	\subsubsection{Deep Lying Posture Tracking}
	
	Recurrent Neural Networks (RNNs) are a type of deep learners that are well-suited to model sequential data. However, RNNs fail to learn long-term dependencies in the data due to the problem of vanishing/exploding gradients. Long Short-term Memory (LSTM) network has been introduced to address this issue and capture long-term dependencies from the sequential time series data  \cite{fang2019performance}. LSTM networks have shown promising results on time series classification tasks \cite{lefebvre2013blstm, graves2013speech}. LSTM captures long-distance dependencies from sequential data through the integration of memory cells and RRNs \cite{gamboa2017deep}. Bidirectional long short-term memory (bi-LSTM) networks were introduced as an extension to the LSTM networks. The bi-LSTM architecture consists of two LSTMs that train in two directions; therefore, it is capable of extracting long-term data dependencies in both forward and backward directions \cite{sun2017bilstm}. \eqref{lstm1} and \eqref{lstm2} show the mathematical formulation of a bidirectional LSTM with $L$ hidden layers. At time-step $t$, each unit in layer $(i)$ receives a set of parameters from the previous time-step (previous state $h_{t-1}^{(i)}$), and two set of parameter from the previous layer (state of previous layer $h_t^{(i-1)}$ and ); one input from the left-to-right RNN and one from the right-to-left RNN. The input to each unit at level $(i)$ is the output of RNN at layer $(i-1)$ at the same time-step $t$. The output, $\widehat{y}$ at each time-step $t$, in \eqref{lstm3}, is the result of propagating the input parameters through all hidden layers.
	
	\begin{equation}
	\label{eq:lstm1}
	    \overrightarrow{h_t}=f(\overrightarrow{W}^{(i)}h_t^{(i-1)}+\overrightarrow{V}^{(i)}\overrightarrow{h}_{t-1}^{(i)}+\overrightarrow{b}^{(i)})
	\end{equation}
	\begin{equation}
	\label{eq:lstm2}
    	 \overleftarrow{h}_t^{(i)}= f(\overleftarrow{W}^{(i)}h_t^{(i-1)}+\overleftarrow{V}^{(i)}\overleftarrow{h}_{t+1}^{(i)}+\overleftarrow{b}^{(i)})
	\end{equation}
	\begin{equation}
	\label{eq:lstm3}
	    \widehat{y}_t=g(Uh_t+c)=g(U[\overrightarrow{h}_t^{(L)};\overleftarrow{h}_t^{(L)}]+c)
	\end{equation}
	
	In the next section, we define lying posture prediction from the sequences of raw sensor data as an optimization problem, then design a deep learning architecture using Bi-LSTM networks as the solution.
	
	\textbf{Problem:}
	We have $N$ sequences of variable lengths where each sequence $X_i$ is assigned a label $Y_i=(y_{i1},y_{i2},...,y{ik})$ using max-likelihood classification, where $y_{ij}$ shows the likelihood of $j_{th}$ class for $i_{th}$ sequence. Given these, the problem is to estimate a set of labels $\hat{Y}=\{\hat{y_{1}}, \hat{y_{2}},..., \hat{y_{N}}\}$, such that the difference between the actual and estimated label sets is minimized. We compute this difference as the cross-entropy of the estimated labels $\hat{Y}$ and actual label $Y$ for summed over the sequences. 
	
	\begin{equation}
	\label{eq:entropy}
	-\sum_{i=1}^{N}\sum_{j=1}^{K}y_{ij}\log{\hat{y_{ij}}}
	\end{equation}
	
	\noindent where $N$ is the number of the samples, $K$ is the number of the classes (e.g., three lying postures). Symbols $y_{ij}$ and $\hat{y_{ij}}$ are the actual and estimated likelihood of $j_{th}$ class for $i_{th}$ sequence, respectively.
	
	Equation \ref{eq:entropy} is not an accurate representation of the error as the sequences might adopt different lengths. Therefore, given a set of scalar numbers $M=\{m_1, m_2, ..., m_i\}$ where $m_i$ is the length of the sequence $X_i$, Equation \ref{eq:entropy} can be formulated as below. 
	
	\begin{equation}
	\label{eq:objective}
	-\sum_{i=1}^{N}\sum_{j=1}^{K}m_iy_{ij}\log{\hat{y_{ij}}}
	\end{equation}
	\noindent where the error is a weighted (e.g., length of sequences) sum of the cross-entropy between the actual and estimated labels. The objective of the sequence classification model is to minimize Equation \ref{eq:objective}.
	
	\textbf{Deep Learning Architecture:}
	To solve this problem, We design an Adaptive LSTM (AdaLSTM), an LSTM Network with an adaptive learning rate method with a decaying learning rate schedule. AdaLSTM receives the sequences of raw accelerometer sensor data as the input and estimates one label for each sequence. As shown in \figref{model}, the input episodes/sequences of raw accelerometer data are fed to a Bi-LSTM layer with ten hidden units. The training process of the bi-LSTM includes back-propagation processes in two directions with the objective of minimizing the error. Three fully connected layers multiply the output of the Bi-LSTM layer (e.g., a sequence of tri-axial accelerometer data) by the matrix of weights and add it by the vector of bias. The output of the fully connected layer is fed to a softmax layer that is a multi-class generalization of the logistic sigmoid function. We compute the cross-entropy loss for multi-class classification based on the likelihood of the softmax function. We set the maximum number of the epochs equal to $100$. We set the initial learning rate and decay rate of the squared gradient moving average to $0.01$ and $0.99$, respectively. To shorten the amount of padding in the mini-batches and make the training more suitable for CPU, we chose the mini-batches to be short sequences of size $27$.  \textit{Adam} optimizer \cite{kingma2014adam} is used for training the neural network through backpropagation.
	
	\section{Experimental Evaluation}
	\subsection{Datasets \& Preprocessing}
	We perform training and validation of the models on two publicly available datasets. (1)  Class-Act dataset \cite{olguin2006human} which contains three major lying postures including supine, prone, and right side from 22 users, and (2) Integration of the Class Act and Daily \& Sport Activities Dataset (DAS) \cite{altun2010comparative} which contain 4 major lying postures including supine, prone, left side, and right side from 8 users.
	
	\subsubsection{Class-Act: Datasets from a Human Posture/Activity classification}
\begin{figure*}[ht]
		\centering
		\begin{subfigure}{.35\textwidth}
			\includegraphics[width=\linewidth]{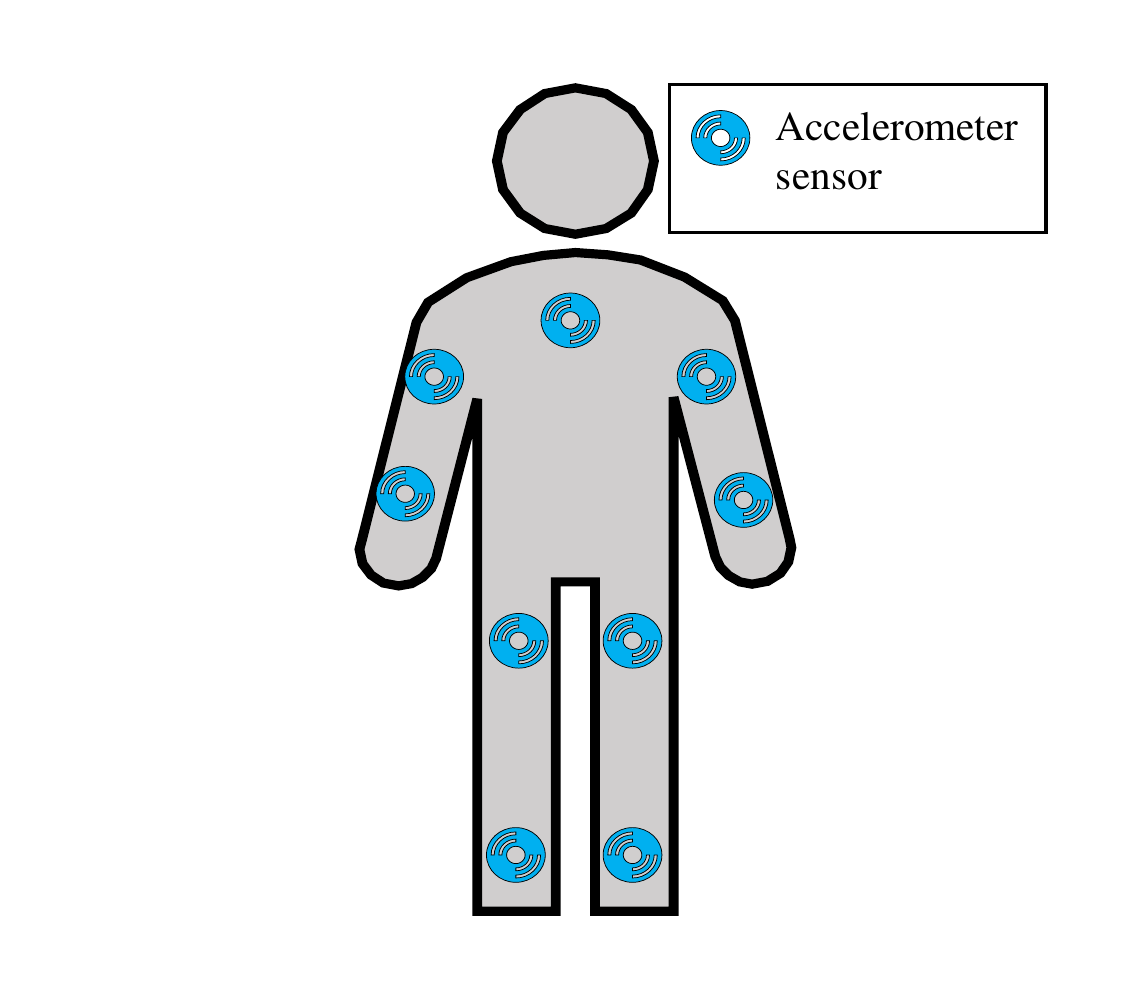}
			\caption{Sensor Positioning}
			\label{fig:sensorpos}
		\end{subfigure}
		\begin{subfigure}{.35\textwidth}
			\includegraphics[width=\linewidth]{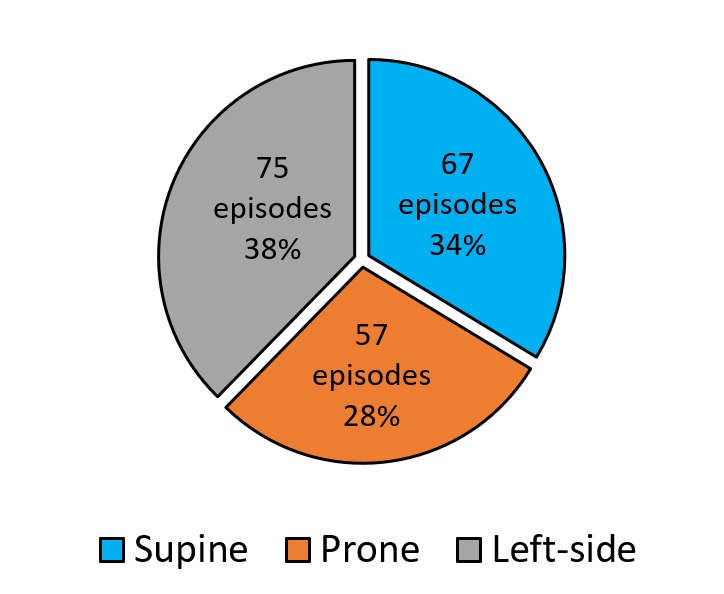}
			\caption{Label Prevalence}
			\label{fig:prevalence}
		\end{subfigure}%
		
		\caption{(a) Visualization of accelerometer sensor positioning, and (b) activity prevalence for Class-Act dataset}
		\label{fig:classact}
	\end{figure*}

	Class-Act is a human posture and activity classification dataset from 22 healthy participants (7 females and 15 males, ages between 20 and 36) \cite{olguin2006human}. The participants worn nine accelerometer sensors sampled at 30 Hz on nine different body locations, including the chest, left and right thigh, left and right ankle, left and right arm, left and right wrist during the activities, as shown in \figref{sensorpos}. Class-Act was collected based on three pre-defined protocols with different combinations of activities or postures in a controlled manner. The activities were walking, sitting, standing, lying supine, lying prone, lying on the left side, kneeling, and crawling. \figref{prevalence} shows the prevalence of the extracted episodes for only lying postures. The duration of different episodes for lying supine, prone, and on the left side were 12.2$\pm$3.6, 11.9$\pm$3.6, and 12.10$\pm$3.3 seconds, respectively.
	
	\subsubsection{Daily and Sports Activities Dataset (DAS)}
	The DAS dataset contains data from eight subjects (four females and four males, ages between 20 and 30) that performed 19 activities of daily living for five minutes each \cite{altun2010comparative}. The participants wore five inertial sensor units embedding a tri-axial accelerometer on the chest, right and left wrist, and right and left thigh. The sensors were calibrated to acquire data at a sampling frequency of 25 Hz. We only used lying supine and lying on the right side postures in this study. The dataset is a balanced dataset, including 16 episodes of data collected from 8 participants while performing lying supine and lying on one side postures, where there were eight episodes for each lying posture. Each episode has a length of 7500 samples (300 seconds).
	
	\subsubsection{Integrated Dataset}
	
	We combined the lying posture episodes from the Class-Act and DAS datasets for the common sensor locations (i.e., chest, right and left wrists, and right and left thighs). We segmented each episode in the DAS dataset into 15 episodes of 500 samples (i.e. 20 seconds). Since the DAS dataset only contained episodes of lying on the back (i.e., supine) and lying on the right side, we performed under-sampling in the DAS dataset prior to combining the datasets to maintain the balance. The final dataset contains 75 episodes of supine (i.e., 26.3$\%$), 57 episodes of prone (i.e., 20.0$\%$), 73 episodes of left side (i.e., 25.6$\%$), and 80 episodes of right side (i.e., 28.1$\%$) for the chest, right and left wrists, and right and left thighs.

	\subsection{Comparison Metrics and Implementation Details}
	We validated the proposed models based on 10 fold validation and leave-one-subject-out (LOSO) validation to minimize overfitting to a specific subject or a specific pattern of performing a lying posture. We report accuracy, F1-Score, and balanced accuracy as evaluation metrics of the proposed models. We perform the coefficient of variation (CoV) analysis \cite{taborri2017factorization} to compare the proposed algorithms against the state-of-the-art.
	
	Accuracy is defined as the average effectiveness of the classifier over all the class labels.
	
	\begin{equation}
	Accuracy = \frac{\sum_{i=1}^{l}\frac{{TP}_{i}+{TN}_{i}}{{TP}_{i}+{TN}_{i}+{FP}_{i}+{FN}_{i}}}{l}
	\end{equation}
	
	where l is the number of class labels. F1-score is defined as the relations between the data’s positive labels and the classifier predicted labels.

	where l is the number of class labels. Balanced accuracy is defined as the average of the true positives and true negatives for each class label. 
	\begin{equation}
	Balanced Accuracy = \frac{\sum_{i=1}^l\frac{{TP}_i}{P_i} + \sum_{i=1}^l\frac{{TN}_i}{N_i}}{2 \times l}
	\end{equation}
	where l refers to the number of classes in the classification task. For each class $C_i$ (i.e., three lying posture classes), $P_i$ is the number of samples with positive label (i.e., $C_i$), $N_i$ is the number of the samples with negative labels (i.e., not  $C_i$), ${TP}_i$ refers to the samples that are correctly classified as $C_i$, ${FP}_i$ are the samples that are falsely classified as $C_i$, ${TN}_i$ are defined as the samples that are correctly not classified as $C_i$, and ${FN}_i$ are the samples that are falsely not classified as $C_i$ \cite{sokolova2009systematic,brodersen2010balanced}.
	
	\begin{equation}
	F1-Score = \frac{2\times(Recall\times Precision)}{Recall+Precision}
	\end{equation}
	where precision refers to the average agreement of the actual class labels and classifier-predicted labels, and recall is the average effectiveness of the classifier to identify each class label. Precision and recall are computed by the following equations.
	\begin{equation}
	    Precision = \frac{\sum_{i=1}^l\frac{{TP}_i}{{TP}_i+{FP}_i}}{l},
		Recall = \frac{\sum_{i=1}^l\frac{{TP}_i}{{TP}_i+{FN}_i}}{l}
	\end{equation}

	We compute the coefficient of variation (CoV) analysis \cite{taborri2017factorization} for each model over different body locations based on the equation below. 
	\begin{equation}
	CoV = \frac{\sigma}{\mu}
	\end{equation}
	\noindent where $\sigma$ and $\mu$ are respectively the standard deviation and average of the F1-Score in lying posture detection over different folds.
	
	\section{Results}
	In this section, we extensively evaluate the performance of the ensemble tree and AdaLSTM classifiers independently and against each other. We report the validation metrics, including accuracy, F1-score, and balanced accuracy for both classification models. We further perform a coefficient of variation (CoV) analysis to compare the performance of the models against the state-of-the-art.
	
	\subsection{Raw data Inspection}
	Before our main analysis, we investigate the variations in the pattern of the raw accelerometer sensor data across different subjects and different lying postures on the Class-Act dataset. We visually inspect the data by computing the mean and standard deviation of the acceleration data over all the episodes of data collected from different subjects while maintaining different lying postures.
	
	\begin{figure}[ht]
		\centering
		\includegraphics[width=4.5in]{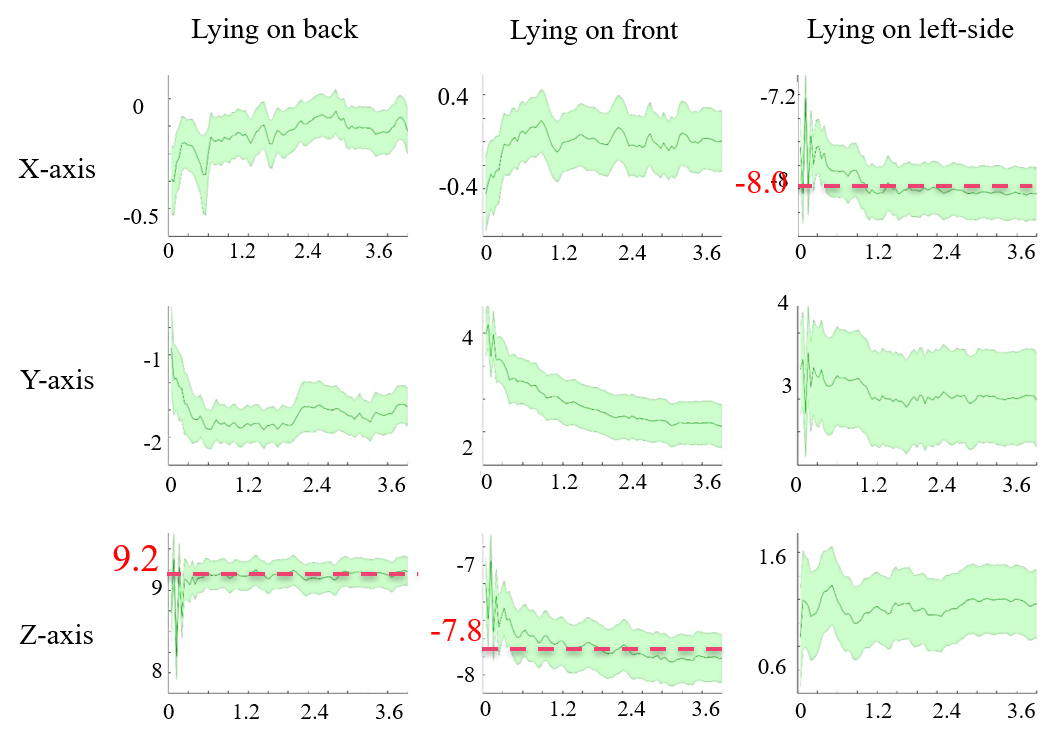}
		\caption{Mean and standard deviation of the magnitude of the accelerometer sensor data for different lying postures over all the subjects.}
		\label{fig:stdmean}
	\end{figure}
	
	These patterns show two issues with the Class-Act dataset: (1) The accelerometer data captured from one of the subjects was not converted to gravity (g) and was stored in analog format. (2) The data collectors labeled multiple episodes of different lying posture as the same posture for one of the subjects. \figref{stdmean} visualizes the changes in 3-axis acceleration for each lying posture on the Class-Act dataset after resolving the issues mentioned above (e.g., correcting sensor output and labels). The solid line demonstrates the mean, and the shaded area shows the standard deviation for different episodes in a specific lying posture. 
	The results show that the y-axis (vertical) always reports values near 0g. In contrast, the x-axis (lateral) shows a mean value near -8.0g for lying on left side posture, while it almost always reports a mean near 0g for the other two postures. Moreover, the z-axis (horizontal) reports mean accelerations around +9.2g and -7.8g for supine and prone postures, respectively, but 0g for lateral posture. Therefore, that lateral axis (x-axis) appears to be sensitive to the lying on side posture, and the horizontal axis (z-axis) appears to be sensitive to the supine and prone postures. 
	These numbers can be justified because while the user is lying on one-side, the x-axis of the accelerometer sensor on the chest is parallel to gravity, therefore reporting values around g ($\pm$10). The same result occurs when the user lies on the back (supine) or front (prone), except the z-axis, becomes parallel to the g and reports values near $\pm$10g. Depending on the direction of the body (supine, prone, and lateral lying postures), these values could be negative or positive. Consequently, we expect these two axes to be more informative in classification compared with the y-axis.
	
	\subsection{Traditional Machine Learning}
	In this section, we validate the feature-based ensemble tree classifier in detecting three major lying postures (supine, prone, and left side) using the Class-Act dataset including 12 subjects and nine sensor locations.
	
	\subsubsection{Feature Engineering} \label{sec:featureselection}
\begin{figure}[ht]
		\includegraphics[width=0.9\textwidth]{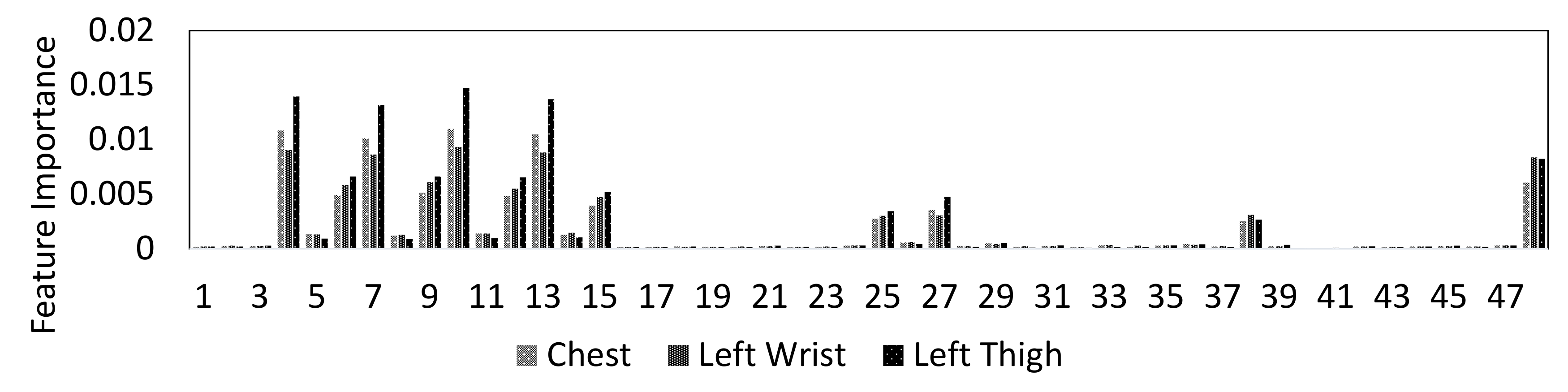}
		\caption{Importance of the extracted features from sensor data for lying posture tracking.}
		\label{fig:features}
	\end{figure}

	\figref{features} shows the feature importance of lying posture tracking as determined by the ensemble tree classifier that is trained on the data from the chest, left thigh and wrist sensors. The importance of the features is one of the outputs of the ensemble tree classification. The y-axis in this figure is the estimation of feature importance from the ensemble tree by summing over the changes in the mean squared error because of splits on every feature and dividing the sum by the number of the branch nodes in the tree \cite{ronao2014human}. The x-axis shows features 1 to feature 48 as in \tblref{features}. Based on the results, features 4, 7, 10, and 13 are the sets with the highest importance. \tblref{features} shows that these features are the median, mean, maximum, and minimum of the vertical axis (e.g., x-axis) of the tri-axial accelerometer sensor. Moreover, features 6, 9, 12, and 15 are the second important set of features. Based on \tblref{features}. These predictors refer to features median, mean, maximum, and minimum of the z-axis of the tri-axial accelerometer signal. These results match the observations on the sensitivity of the lying on the left side to the x-axis and lying supine and lying prone postures to the frontal axis (e.g., z-axis) in \figref{stdmean}.
	
	\subsubsection{Lying Posture Detection}
	
	\tblref{fb_all} reports the average and standard deviation of accuracy, balanced accuracy, and F1-score of lying posture detection using the ensemble tree classifier. Overall, the body locations such as the chest and the left thigh, which are less susceptible to nocturnal movement during sleep, demonstrate high performance in lying posture detection, While, sensor locations such as the arms and the wrists are poor in lying posture detection. In particular, the ensemble tree classifiers trained on the data collected from the chest, thighs or ankles achieve 89.8\% --96.2\% accuracy, 89.8\%--96.2\% balanced accuracy and, 89.8\%--96.2\% F1-Score. While, these values drop when the sensor is placed on the upper body parts such as the arms and the wrists (78.6\%--89.5\% accuracy, 62.9\%--84.0\% balanced accuracy, and 64.1\%--81.6\% F1-Score).

	\begin{table}[ht]
		\centering
		\small
		\caption{Performance (\%) of the ensemble tree classification in lying-posture detection for nine different body locations on the Class-Act dataset.}
	\addtolength{\tabcolsep}{-2pt}
	\renewcommand{\arraystretch}{1.2}
		\begin{tabular}{crrr}
			\hline
			\textbf{Location} 
			& \textbf{Accuracy}     & \begin{tabular}[c]{@{}c@{}}\textbf{Balanced} \\ \textbf{Accuracy} \end{tabular}  & \textbf{F1-Score}     \\ \hline

			\textbf{Left Thigh}& 94.5$\pm$ 6.9 & 91.3$\pm$10.3 & 90.7$\pm$11.8 \\\hline
			
			\textbf{Right Thigh}& 96.2$\pm$ 8.1 & 94.4$\pm$12.0 & 93.5$\pm$14.4 \\\hline
			
			\textbf{Left Ankle}& 94.9$\pm$8.5 & 92.1$\pm$12.8 & 91.4$\pm$15.6 \\\hline
			
			\textbf{Right Ankle}& 89.8$\pm$13.5 & 82.9$\pm$19.9 & 82.8$\pm$22.7\\\hline
			
			\textbf{Chest} &96.2$\pm$9.1 & 93.6$\pm$13.7 & 93.6$\pm$16.2 \\\hline
			
			\textbf{Left Arm}& 78.6$\pm$11.7 & 62.9$\pm$15.1 & 60.9$\pm$16.6\\\hline
			
			\textbf{Right Arm}& 89.5$\pm$12.1 & 84.0$\pm$18.3 & 81.6$\pm$21.7 \\\hline
			
			\textbf{Left Wrist}& 78.6$\pm$12.5 & 67.1$\pm$19.1 & 64.1$\pm$21.7\\\hline
			
			\textbf{Right Wrist}& 80.7$\pm$14.1 & 79.7$\pm$21.3 & 67.9$\pm$23.8 \\\hline
			
		\end{tabular}
		\label{tbl:fb_all}
	\end{table}
	
\begin{figure*}[t]
	\centering
	\begin{subfigure}{.3\linewidth}
	    \includegraphics[width=\linewidth]{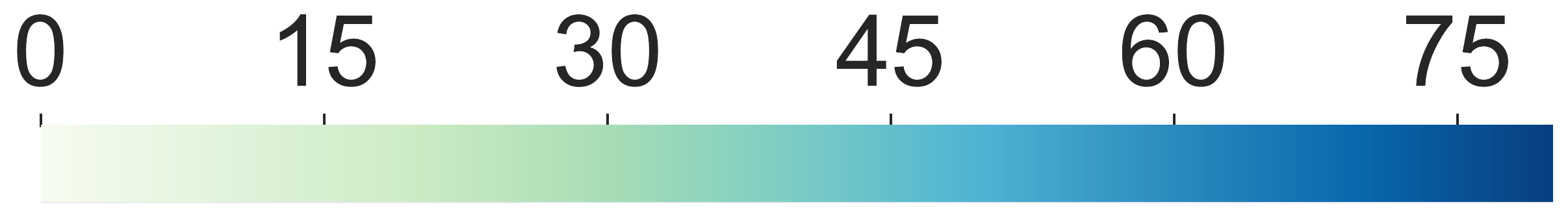}
	\end{subfigure}
	
		\begin{subfigure}{.3\linewidth}
			\includegraphics[width=\linewidth]{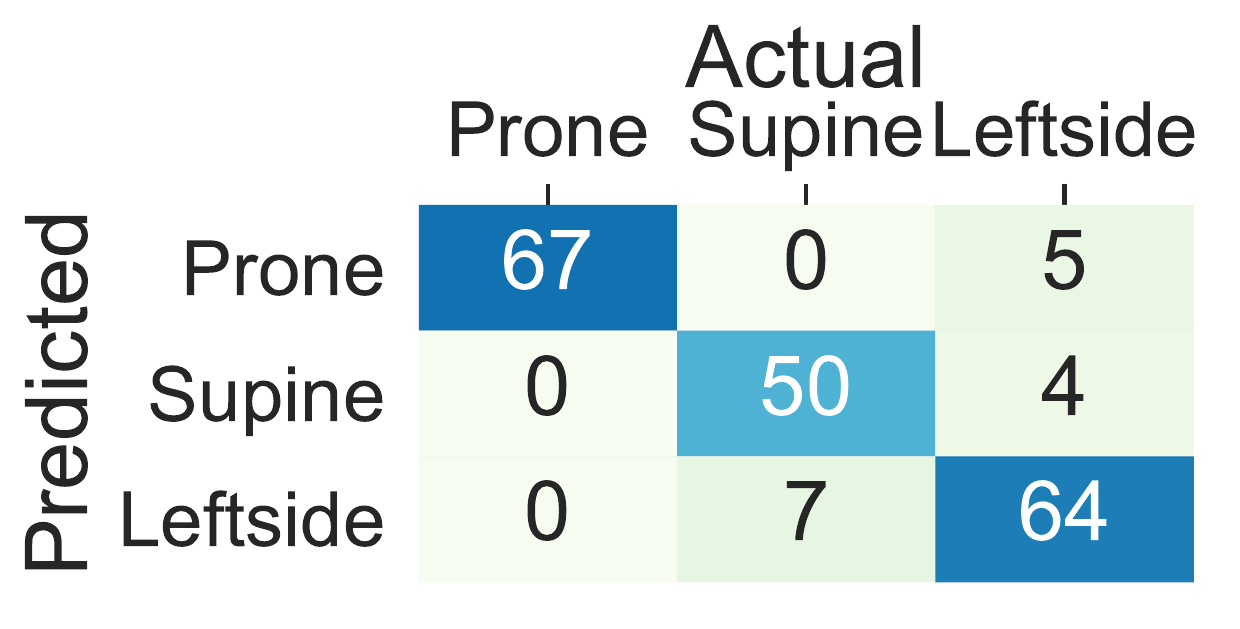}
			\caption{Left thigh}
			\label{fig:et_lt}
		\end{subfigure}
		\begin{subfigure}{.3\textwidth}
			\includegraphics[width=\linewidth]{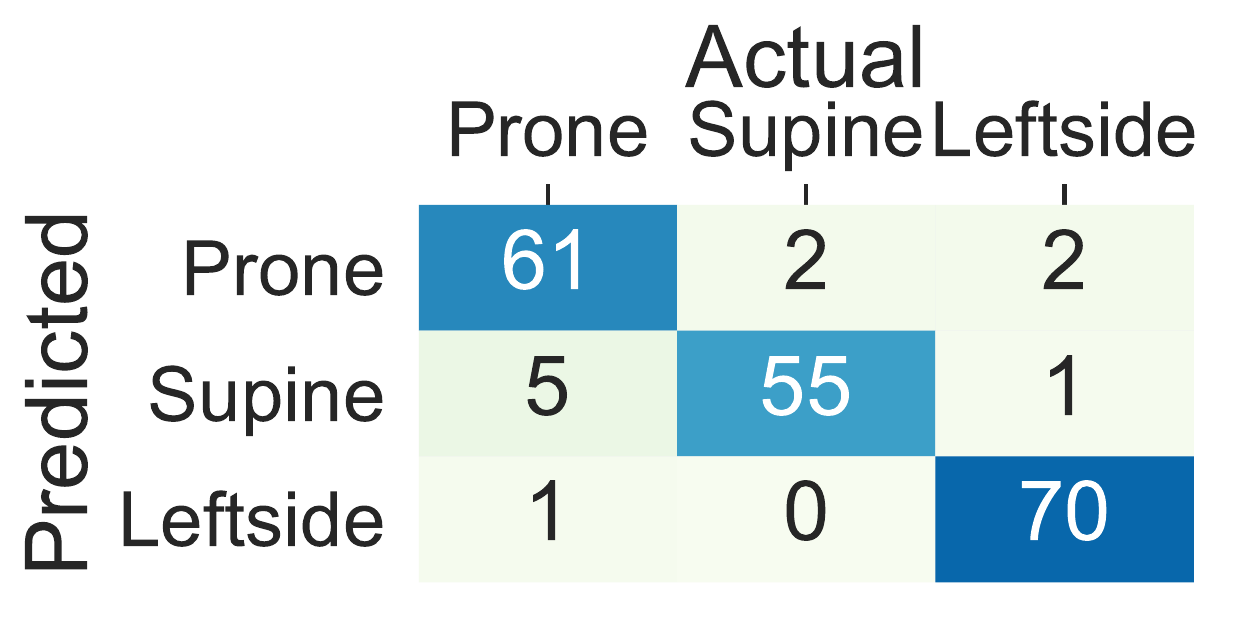}
			\caption{Right thigh}
			\label{fig:et_rt}
		\end{subfigure}
		\begin{subfigure}{.3\textwidth}
			\includegraphics[width=\linewidth]{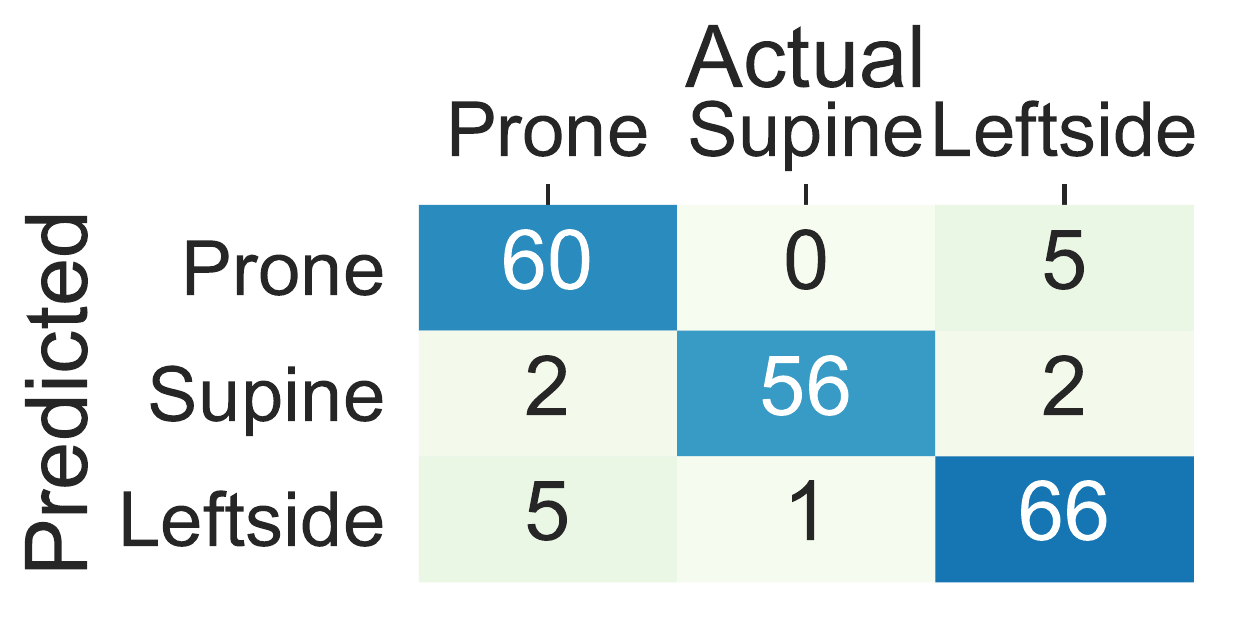}
			\caption{Left Ankle}
			\label{fig:et_ran}
		\end{subfigure}
		\begin{subfigure}{.3\textwidth}
			\includegraphics[width=\linewidth]{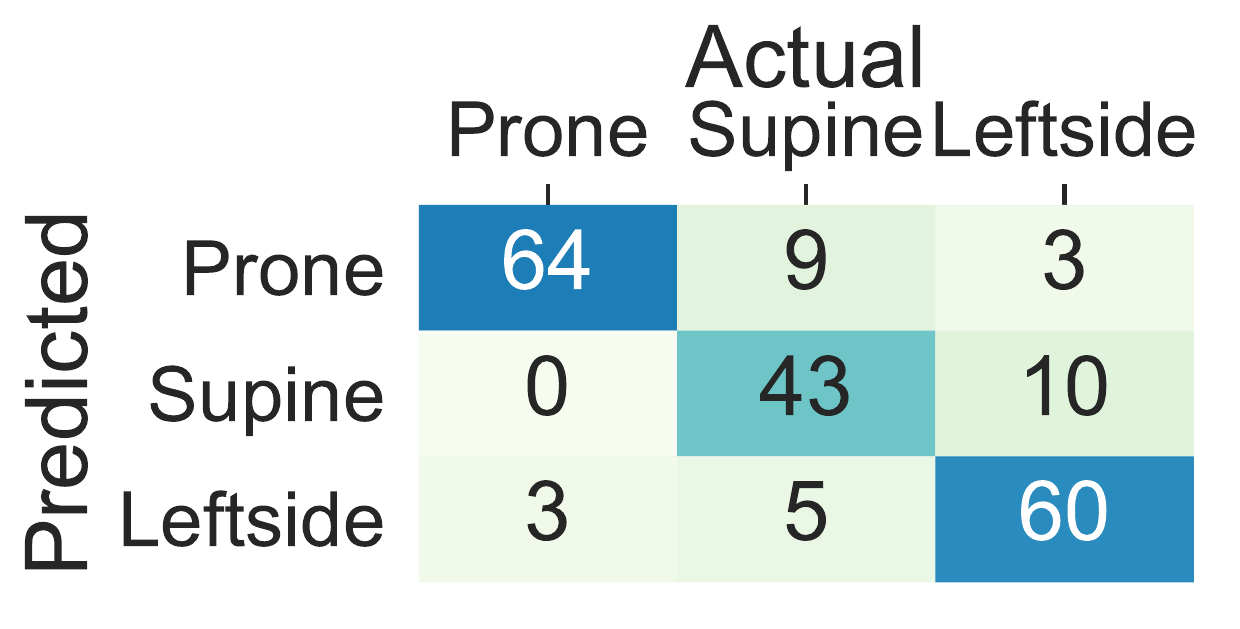}
			\caption{Right Ankle}
			\label{fig:et_ran}
		\end{subfigure}
		\begin{subfigure}{.3\textwidth}
			\includegraphics[width=\linewidth]{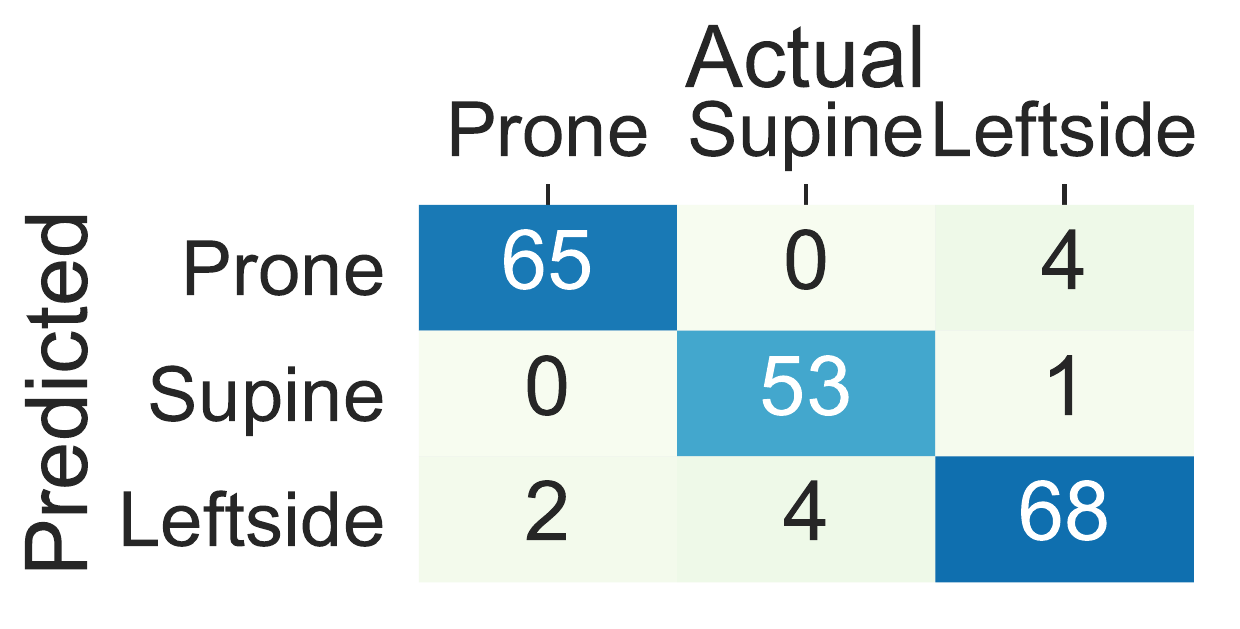}
			\caption{Chest}
			\label{fig:et_c}
		\end{subfigure}
		\begin{subfigure}{.3\textwidth}
			\includegraphics[width=\linewidth]{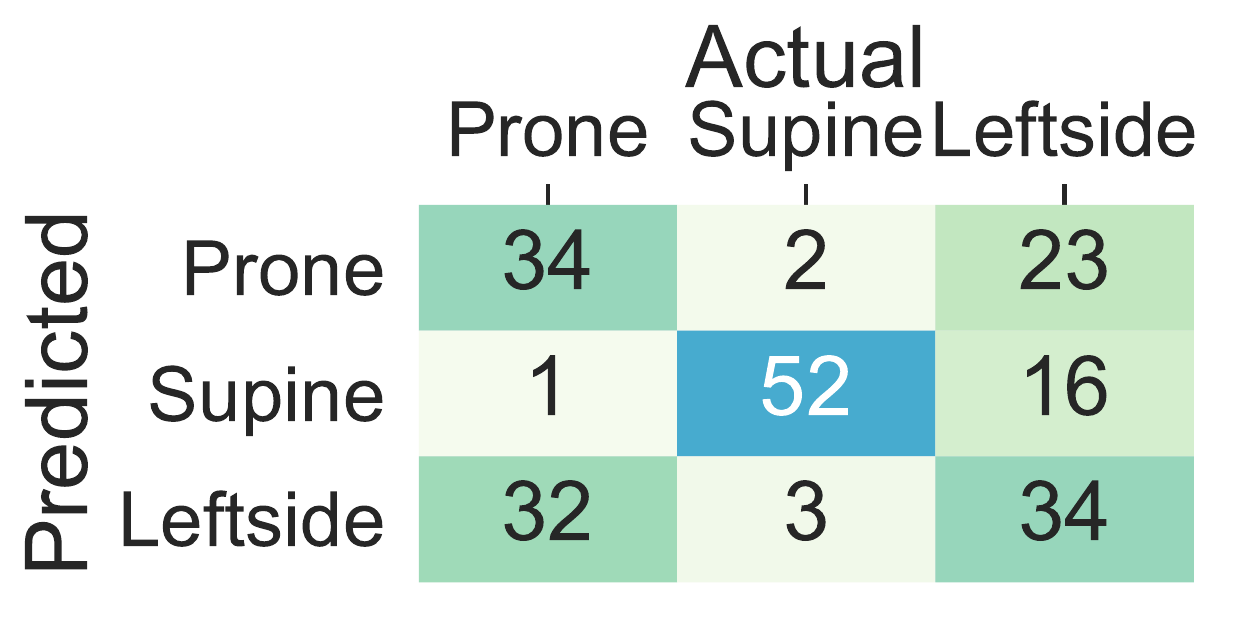}
			\caption{Left arm}
			\label{fig:et_lar}
		\end{subfigure}
		\begin{subfigure}{.3\textwidth}
			\includegraphics[width=\linewidth]{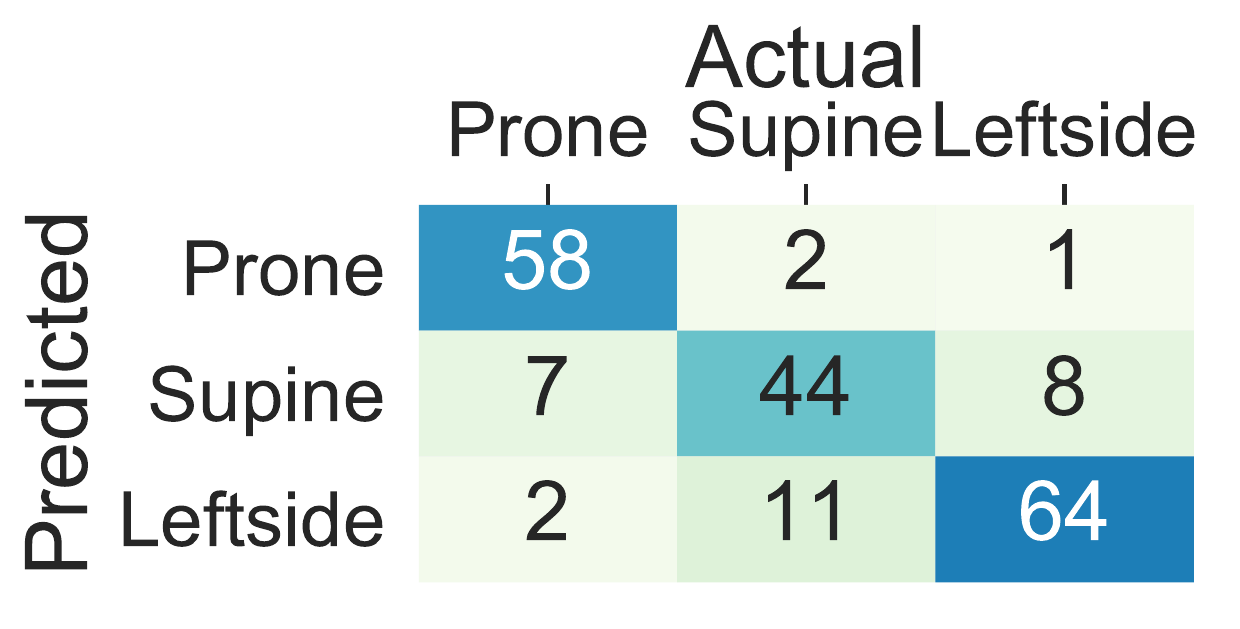}
			\caption{Right arm}
			\label{fig:et_rar}
		\end{subfigure}
		\begin{subfigure}{.3\textwidth}
			\includegraphics[width=\linewidth]{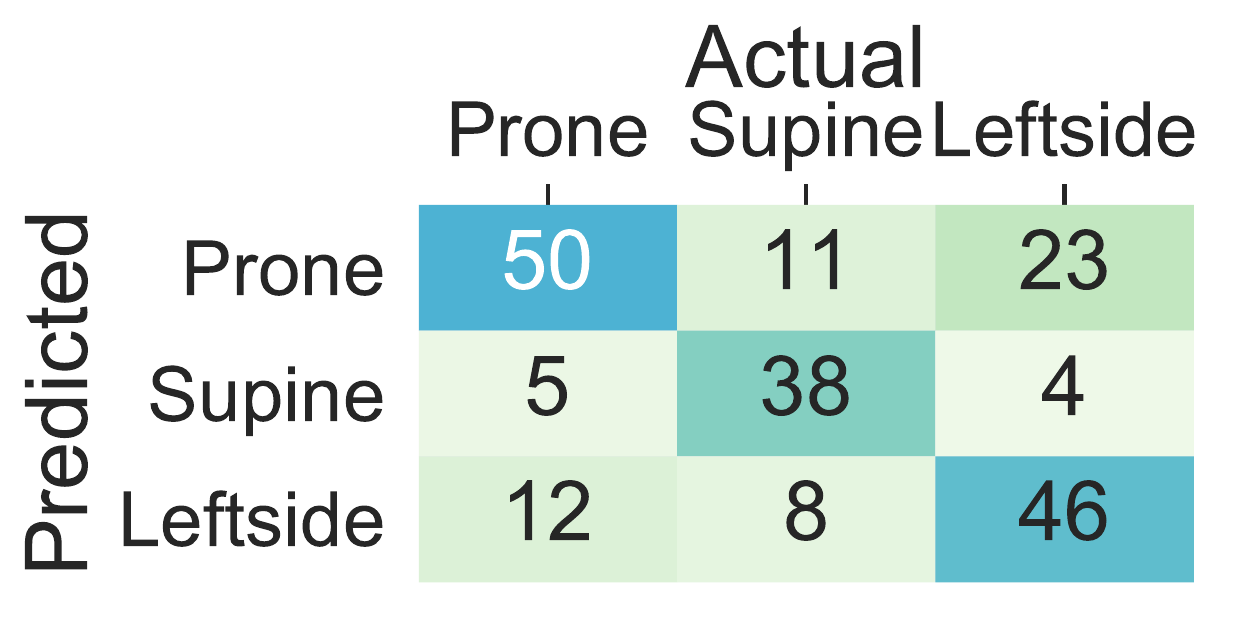}
			\caption{Left Wrist}
			\label{fig:et_lw}
		\end{subfigure}
		\begin{subfigure}{.3\textwidth}
			\includegraphics[width=\linewidth]{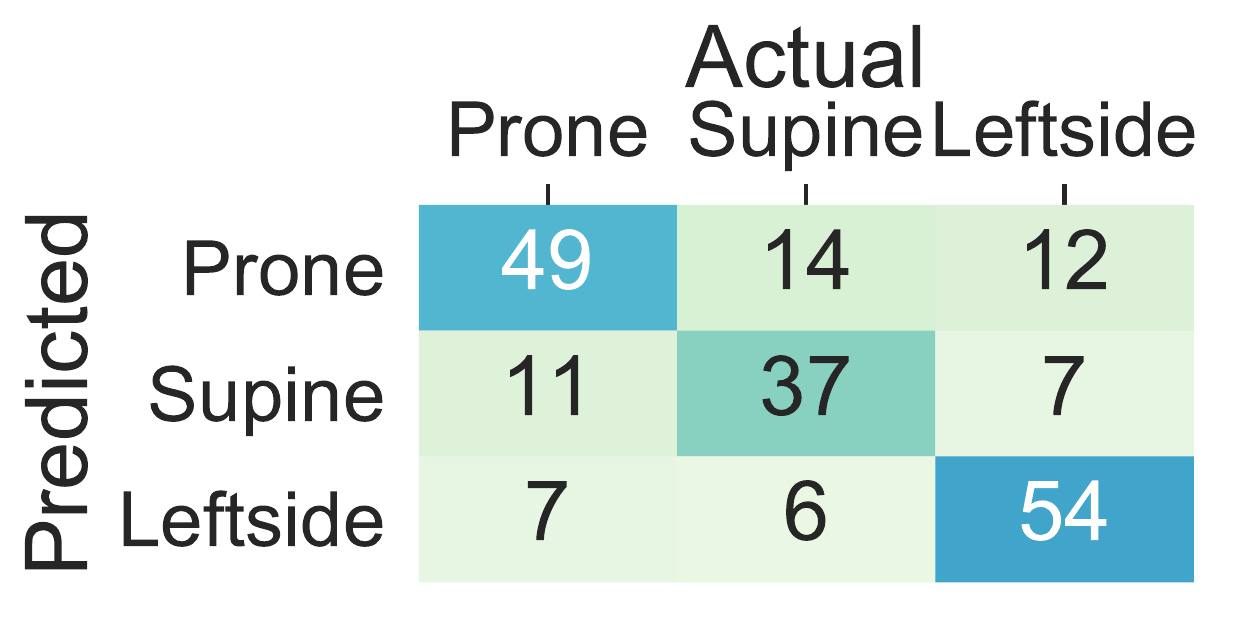}
			\caption{Right wrist}
			\label{fig:et_rw}
		\end{subfigure}
		\caption{Confusion matrix of the ensemble tree classifier in classifying lying postures into supine, prone, and left side for the thighs, ankles, chest, arms, and wrists locations. }
		\label{fig:conf_et}
	\end{figure*}

	The performance decline in the upper body parts is originated from inter-subject variations in the placement of the arms and the wrists and nocturnal movements of them such as bending and rotating while maintaining the same lying posture. The trend in the standard deviation of accuracy, balanced accuracy, and F1-Score across different subjects is also in concordance with the hypothesis that more within-subject variation (e.g., $9.2\%$ to $26.7\%$ standard deviation in accuracy, balanced accuracy, and F1-Score) is observed when the sensor is worn on the wrists and the arms comparing to the chest, thighs and ankles (e.g., $6.2\%$ to $15.6\%$ standard deviation in accuracy, balanced accuracy and F1-Score). Moreover, \figref{conf_et} visualizes the confusion matrix of lying posture detection training the ensemble tree classifier using the Class-Act dataset. The confusion matrices of the classifying the thighs, chest, and ankles data show more promising results than the arms and wrists. In particular, the classifiers trained on the left thigh, right thigh, and chest misclassify 5.5\%, 8.1\%, 5.5\%, and 7.6\% of the lying episodes. The misclassification rate increases to 15.2\% and 15.7\% for the left ankle and left arm locations, and 39.1\%, 31.9\%, and 28.93\% for the right ankle, right arm, and right wrist locations.

	\subsection{Deep Sequence Learning}
	    In this section, we evaluate the performance of the AdaLSTM classifier in detecting three major lying postures. \tblref{lstm_all} shows the mean and standard deviation of average accuracy, balanced accuracy, and F1-score of the model using the Class-Act dataset including nine sensor locations, and 12 subjects.  AdaLSTM achieves 89.8\%--96.2\% average Accuracy, 89.8\%--96.2\% average balanced accuracy, and 89.8\%--96.2\% average F1-Score when applied to the data collected from the sensor worn on the chest, thighs, or ankles. However, the performance drops to 78.6\%--89.5\% average accuracy, 67.1\%--84.0\% average balanced accuracy, and 64.1\%--81.6\% average F1-Score, for the cases of the sensor on the arms and wrists. The within-subject standard deviation in the accuracy, balanced accuracy, and F1-Score is higher when the sensor is placed on the arms and the wrists (11.7\%--23.8\%) comparing to the thighs and chest (6.9\%--13.7\%). Such results could be justified according to the findings in a study by Skarpsno et al. were showed the duration of nocturnal movements while sleeping in the arms and upper back was higher than the thighs in 2,107 subjects \cite{skarpsno2017sleep}. AdaLSTM is the most accurate when applied to the data collected from the sensor on the right thigh (96.2\% $\pm$ 8.1 accuracy, 96.2\% $\pm$ 8.1 balanced accuracy, and 96.2\% $\pm$ 8.1 F1-score). The model on the left wrist achieves the lower performance such as 78.6\% $\pm$ 12.5\% accuracy, 67.1\% $\pm$ 19.1\% balanced accuracy, and 64.1\% $\pm$ 21.7\% F1-Score.
	
	\begin{table}[ht]
		\small
		\caption{Performance (\%) of the sequence classification using AdaLSTM in lying-posture detection for nine different body locations on the Class-Act dataset.}
		\centering
		\addtolength{\tabcolsep}{-2pt}
		\renewcommand{\arraystretch}{1.2}
		\begin{tabular}{crrr}
			\hline
            \textbf{Location}
			& \textbf{Accuracy}     & \begin{tabular}[c]{@{}c@{}}\textbf{Balanced} \\ \textbf{Accuracy} \end{tabular}  & \textbf{F1-Score} \\ \hline

			\textbf{Left Thigh}  & 98.9$\pm$8.2 & 98.4$\pm$5.2      & 98.2$\pm$6.2 \\\hline

			\textbf{Right Thigh}   & 95.9$\pm$7.3 & 93.4$\pm$11.8      & 91.5$\pm$15.6  \\\hline
			
			\textbf{Left Ankle}  & 97.9$\pm$4.2 & 96.8$\pm$6.3 & 96.9$\pm$6.4  \\\hline
			
			\textbf{Right Ankle}   & 94.5$\pm$6.3 & 92.4$\pm$9.4 & 91.7$\pm$10.7 \\\hline
			
			\textbf{Chest}    & 98.3$\pm$7.1     & 97.4$\pm$7.1      & 97.3$\pm$7.3  \\\hline
			
			\textbf{Left Arm}  & 77.6$\pm$11.7 & 68.8$\pm$14.1      & 66.3$\pm$16.5  \\\hline
			
			\textbf{Right Arm}   & 86.8$\pm$9.2 & 79.0$\pm$14.5      & 75.7$\pm$17.3  \\\hline
			
			\textbf{Left Wrist}  & 64.8$\pm$22.9 & 64.9$\pm$24.8 & 62.9$\pm$23.2  \\\hline
			
			\textbf{Right Wrist} & 66.8$\pm$26.7 & 67.6$\pm$26.2  & 66.9$\pm$28.9  \\\hline
		\end{tabular}
		\label{tbl:lstm_all}
	\end{table}
	
\begin{figure*}[ht]
	\centering
	\begin{subfigure}{.3\linewidth}
	    \includegraphics[width=\linewidth]{figures/cbar.pdf}
	\end{subfigure}
	
		\begin{subfigure}{.3\linewidth}
			\includegraphics[width=\linewidth]{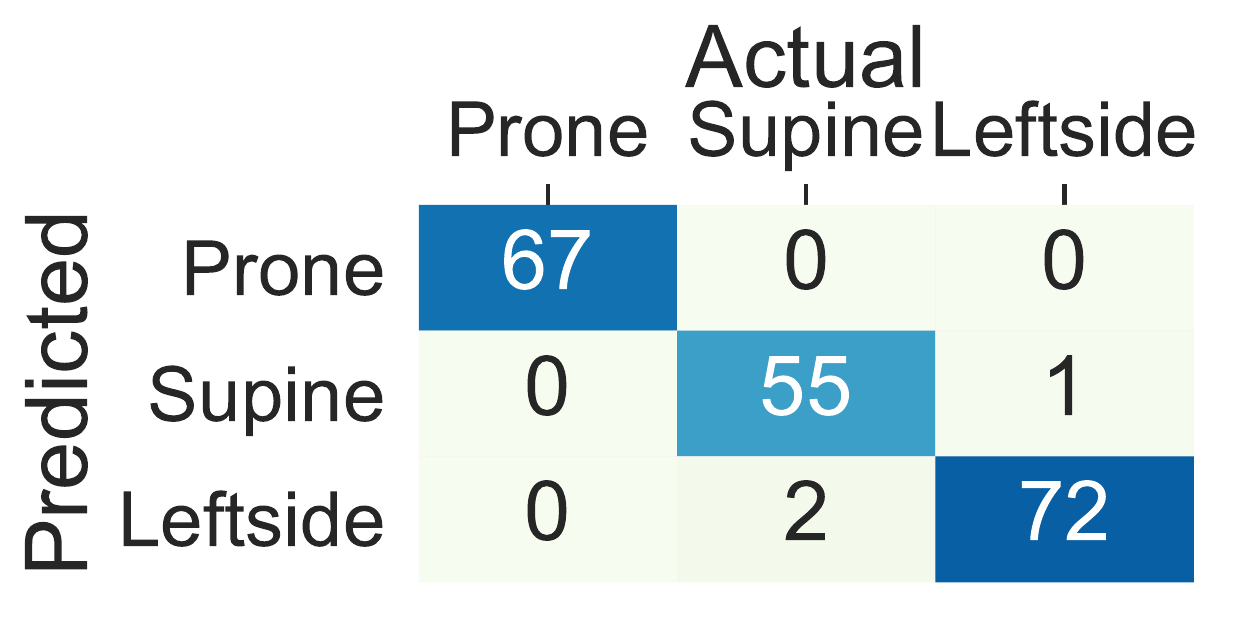}
			\caption{Left thigh}
			\label{fig:et_lt}
		\end{subfigure}
		\begin{subfigure}{.3\textwidth}
			\includegraphics[width=\linewidth]{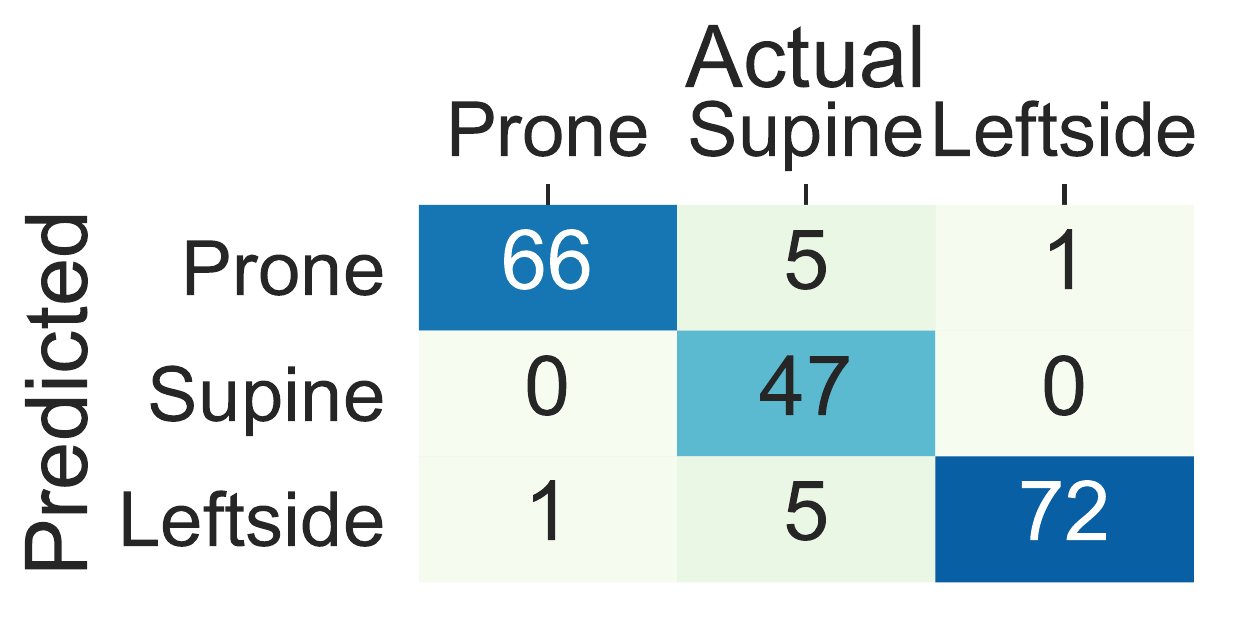}
			\caption{Right thigh}
			\label{fig:et_rt}
		\end{subfigure}
		\begin{subfigure}{.3\textwidth}
			\includegraphics[width=\linewidth]{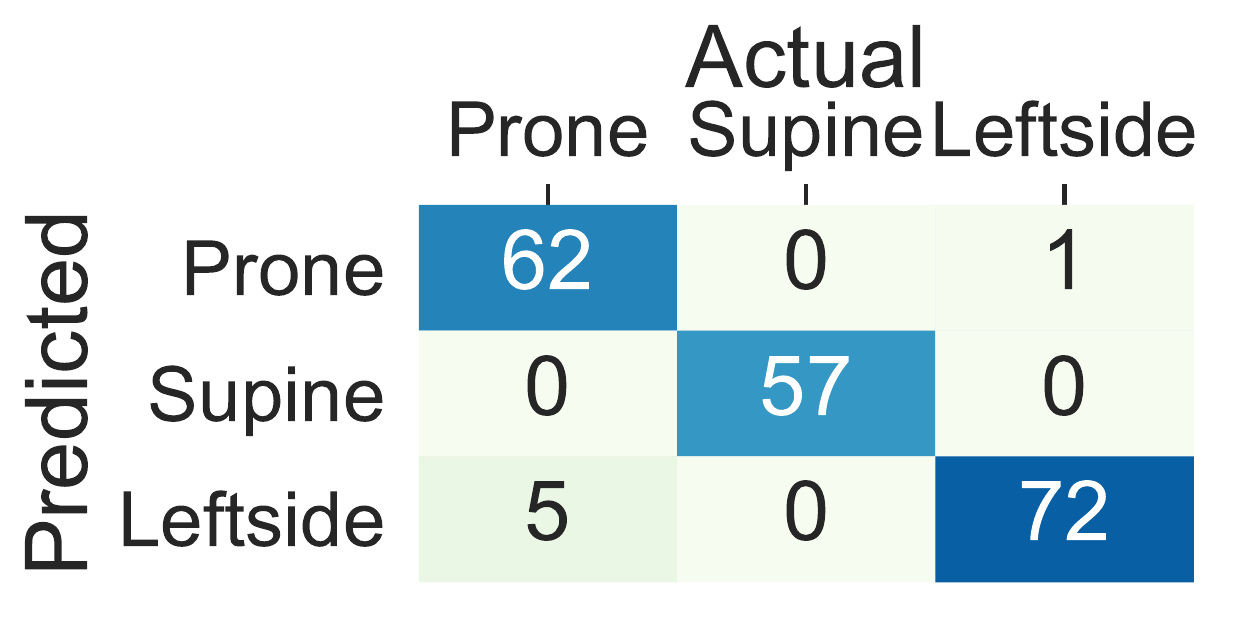}
			\caption{Left Ankle}
			\label{fig:et_ran}
		\end{subfigure}
		\begin{subfigure}{.3\textwidth}
			\includegraphics[width=\linewidth]{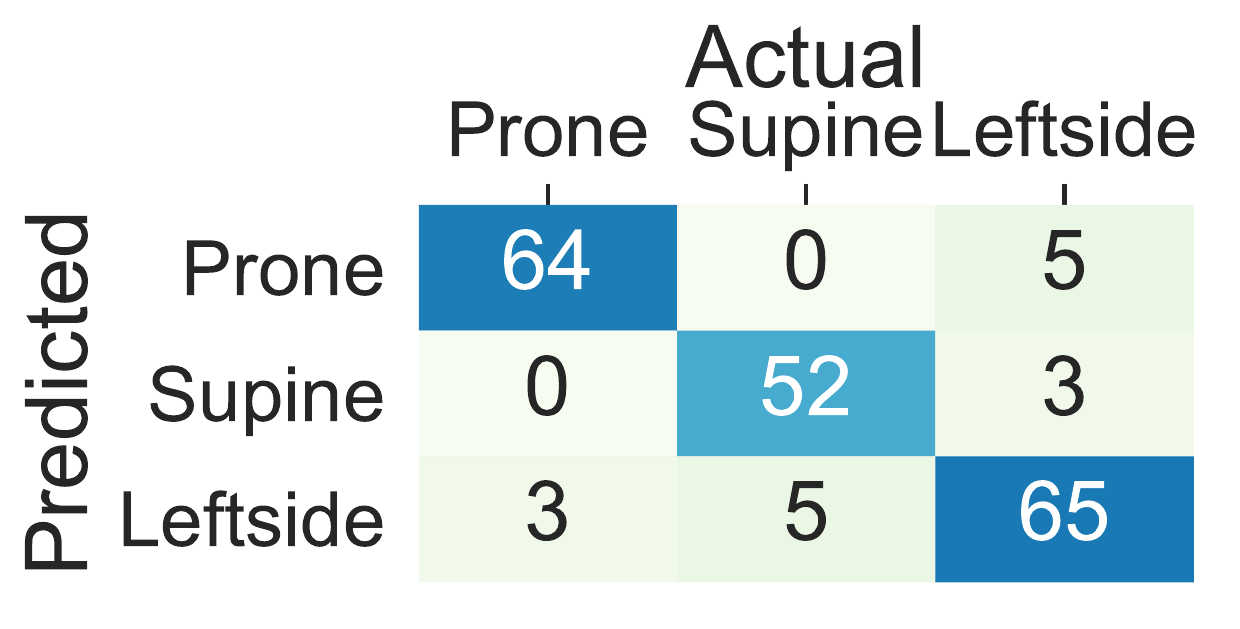}
			\caption{Right Ankle}
			\label{fig:et_ran}
		\end{subfigure}
		\begin{subfigure}{.3\textwidth}
			\includegraphics[width=\linewidth]{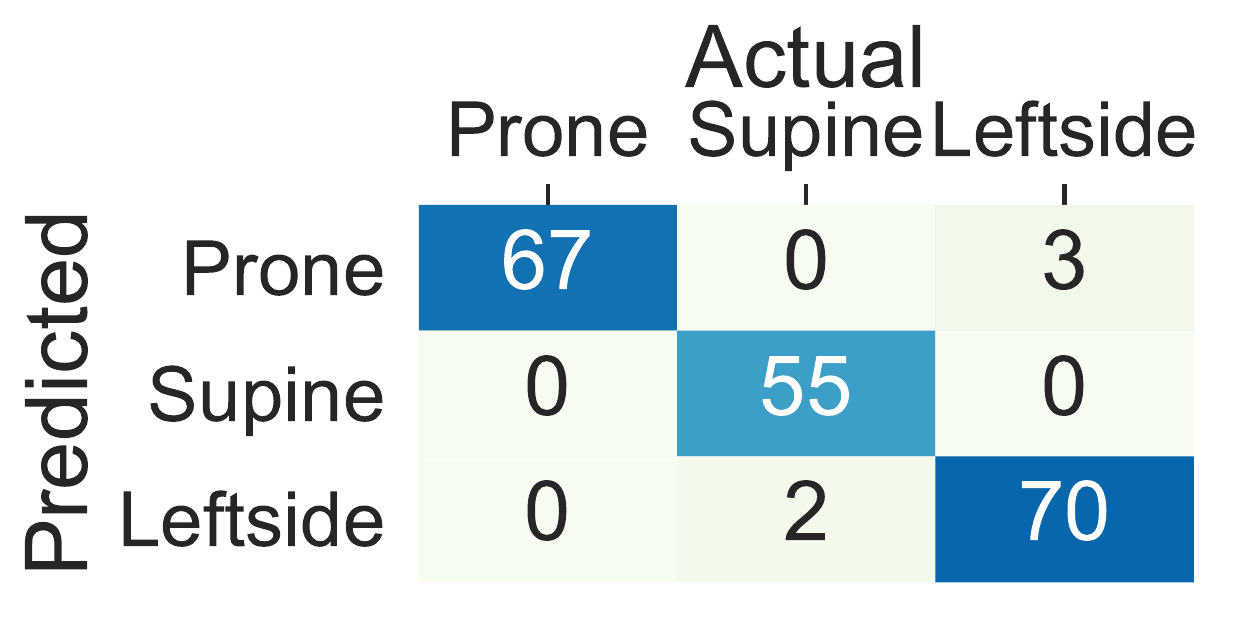}
			\caption{Chest}
			\label{fig:et_c}
		\end{subfigure}
		\begin{subfigure}{.3\textwidth}
			\includegraphics[width=\linewidth]{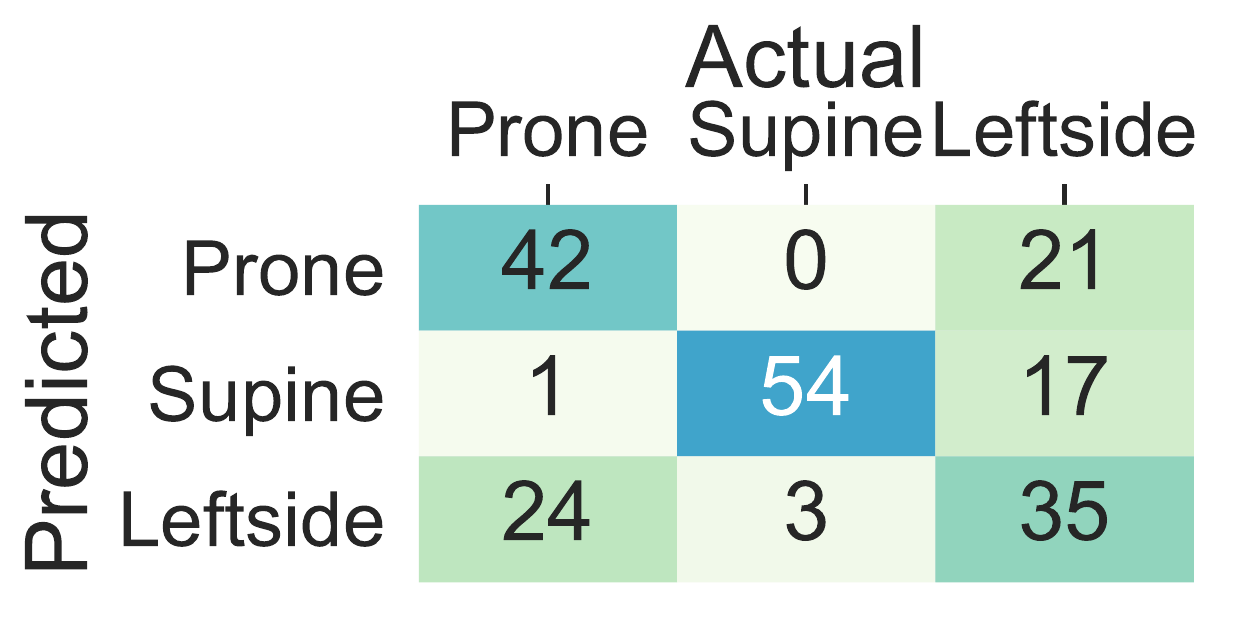}
			\caption{Left arm}
			\label{fig:et_lar}
		\end{subfigure}
		\begin{subfigure}{.3\textwidth}
			\includegraphics[width=\linewidth]{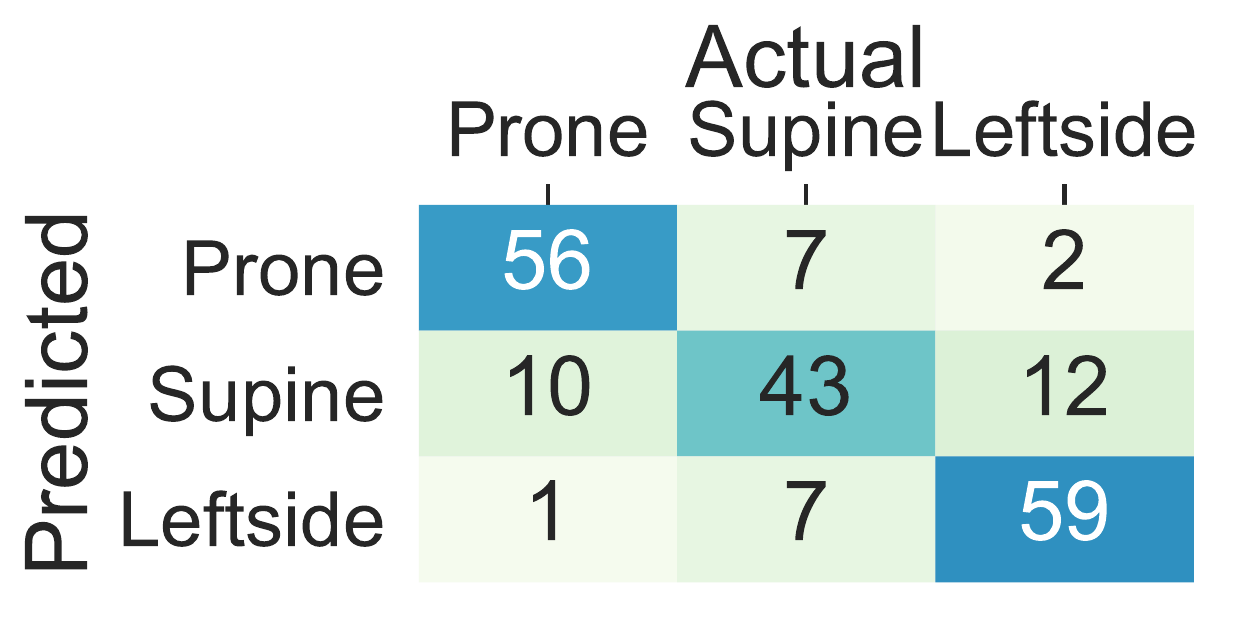}
			\caption{Right arm}
			\label{fig:et_rar}
		\end{subfigure}
		\begin{subfigure}{.3\textwidth}
			\includegraphics[width=\linewidth]{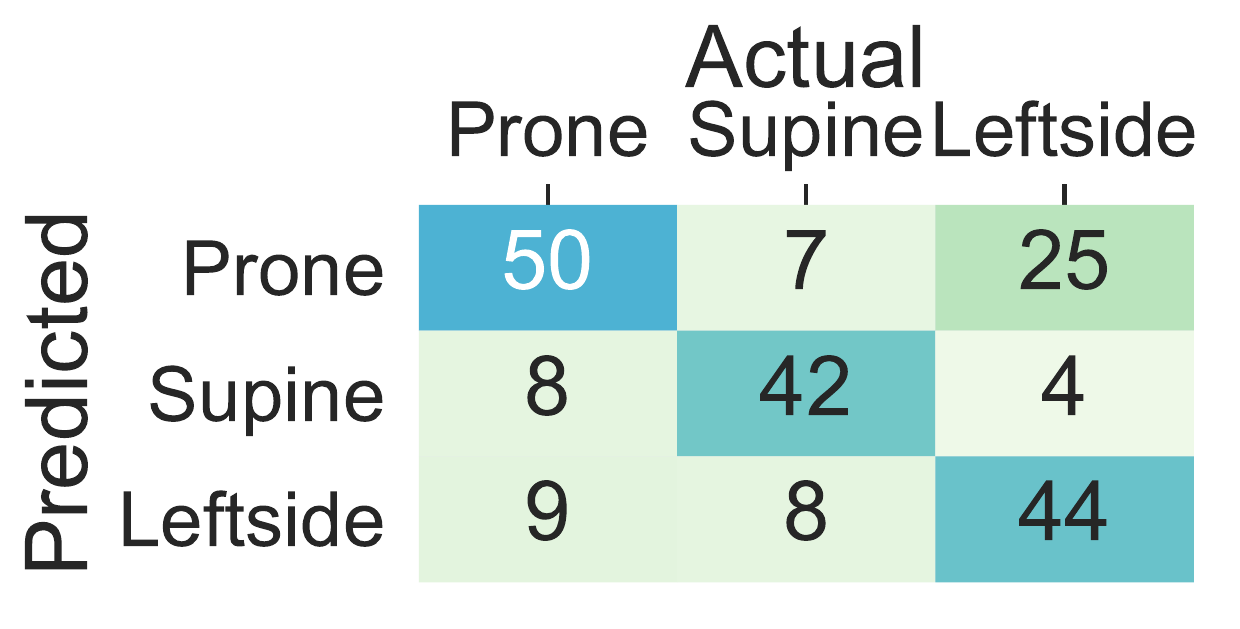}
			\caption{Left Wrist}
			\label{fig:et_lw}
		\end{subfigure}
		\begin{subfigure}{.3\textwidth}
			\includegraphics[width=\linewidth]{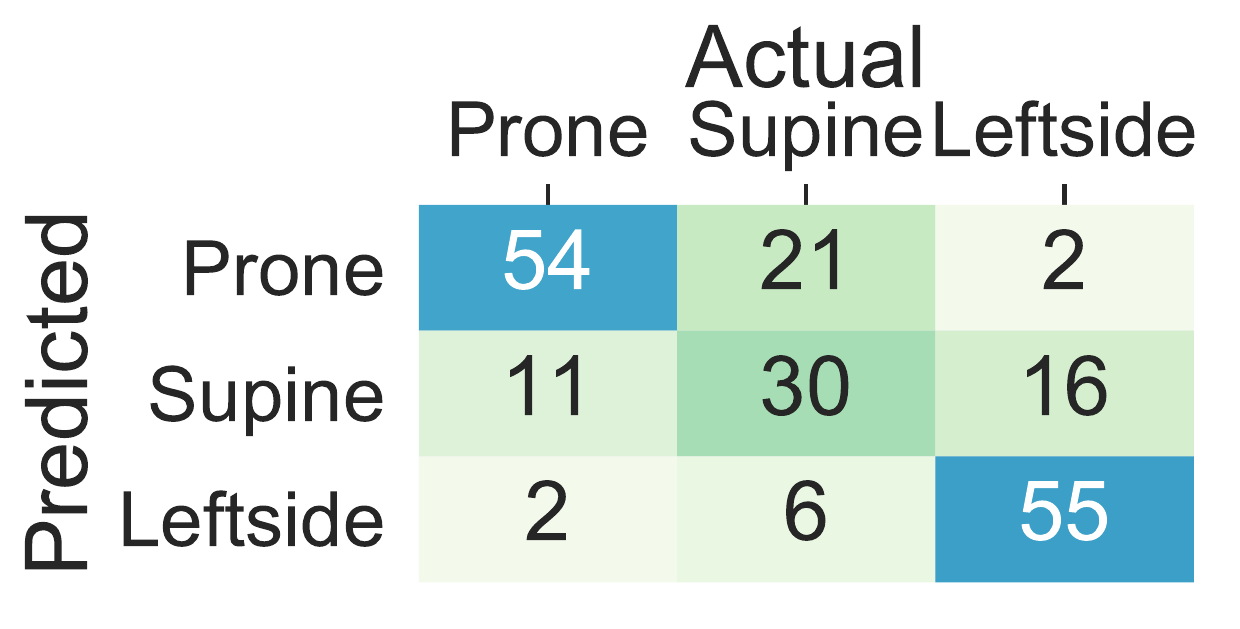}
			\caption{Right wrist}
			\label{fig:et_rw}
		\end{subfigure}
		\caption{Confusion matrix of AdaLSTM classifier in classifying lying postures into supine, prone, and left side for the thighs, ankles, arms, and wrists locations.}
		\label{fig:conf_lstm}
	\end{figure*}

	\figref{conf_lstm} shows the confusion matrices of the AdaLSTM classifier trained on the data from the thighs, ankles, arms, and wrists using Class-Act dataset. The models trained on the chest, left thigh, right thigh, left ankle, and right ankle confuses 2.5\% 1.5\%, 6.1\%, 3.0\%, and 8.1\% of the lying episodes. However, the number of misclassified episodes increases to 33.5\%, 19.8\%, 30.9\%, and 29.4\% for the left arm, right arm, left wrist, and right wrist classifiers. We note that the higher misclassification rate when the sensor is on the left arm than the right arm sensor is due to the confusion of the left side and prone postures.

	\subsection{Deep Learning vs. Traditional Machine Learning}
	
\begin{figure}[ht]
	\begin{subfigure}{\textwidth}
		\includegraphics[width=\linewidth]{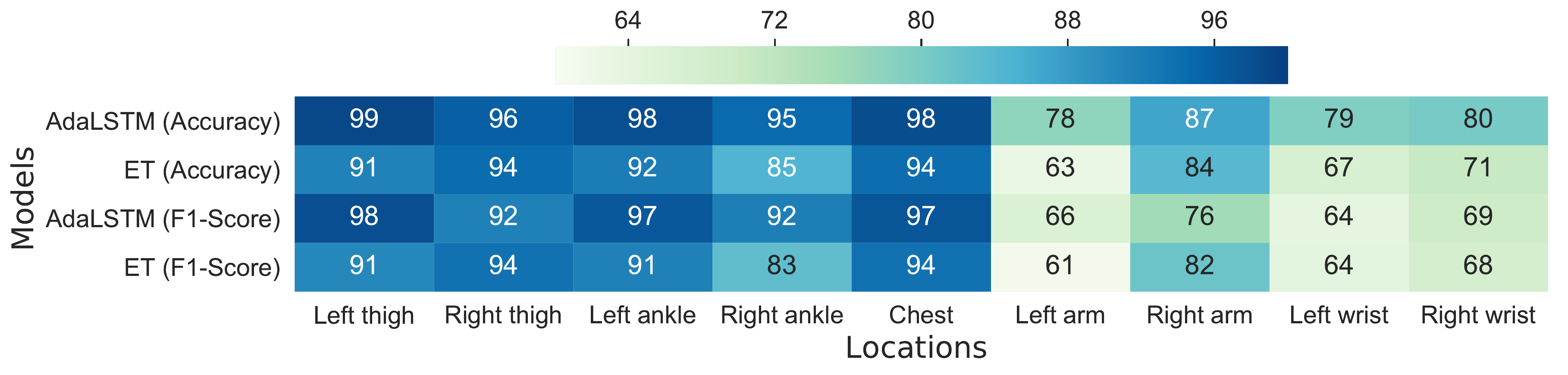}
			\caption{Mean Value}
			\label{fig:mean}
		\end{subfigure}
		\begin{subfigure}{\textwidth}
			\includegraphics[width=\linewidth]{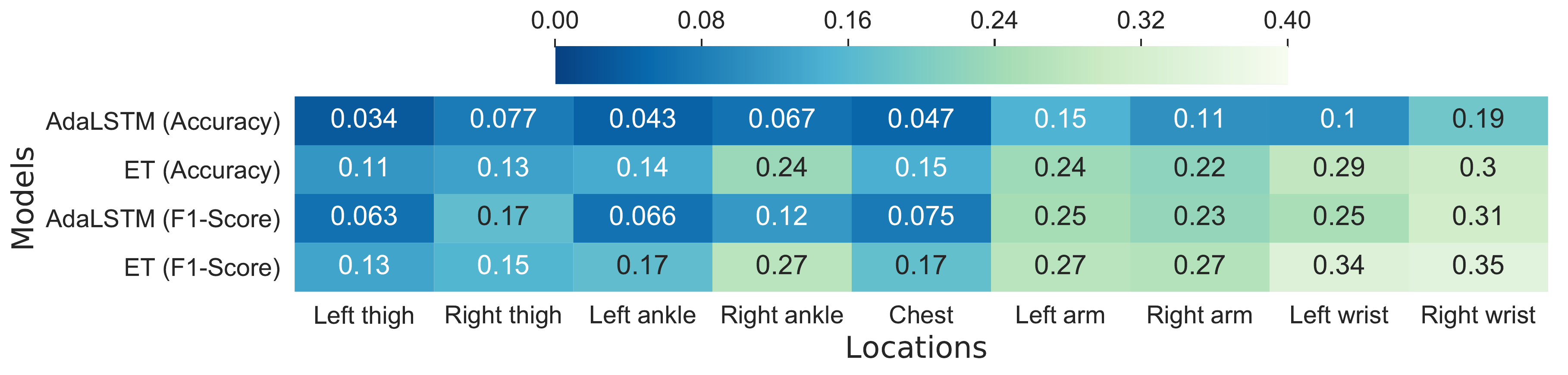}
			\caption{CoV}
			\label{fig:cov}
		\end{subfigure}%
		\caption{Comparison between the mean and CoV of F1-Score (\%) of the ensemble tree and AdaLSTM classification models for nine body locations on the Class-Act dataset using LOSO validation.}
		\label{fig:fb_lstm}
	\end{figure}

	In this section, we investigate the possibility of replacing feature-based machine learning models with deep recurrent neural networks (RNNs). \figref{fb_lstm} compares the mean and CoV of F1-Score and accuracy metrics between AdaLSTM and ensemble tree classifiers. These classifiers are evaluated using the Class-Act dataset including 12 subjects and nine sensor locations on the body. As shown in \figref{mean}, AdaLSTM achieves 2\%--10\% higher accuracy and 3\%--9\% higher F1-Score than Ensemble tree classifier when applied to the data collected from the sensor on the chest, the thighs, or the ankles. The gap between the performance of the two classifiers increases to 3\%--15\% in accuracy when tested on the data collected from the arms or the wrists. Since CoV represents the ratio of variation to the mean of a metric, lower CoV values show a more promising classification performance. As shown in \figref{cov}, AdaLSTM achieves lower CoV values over all the sensor locations, therefore, it adopts a better generalization to cross-subject variations comparing to the ensemble tree classifier. This gap between the performance of the models demonstrates that deep RNNs are more capable of capturing higher-level patterns in noisy data with high variance across subjects such as the data collected from the sensor on the wrists, or the arms.
	
	In addition, we performed Kruskal's statistical test between the CoV values of the AdaLSTM and ensemble tree classifiers to identify any significant difference between the median of the 
	two groups. The Kruskal's test on the CoV of F1-Score, and accuracy show p-value of 0.100, and 0.006, respectively. These results could not reject the null hypothesis, therefore, show no significant difference between the performance of the two classifiers. Note that both 0.100, and 0.007 are marginally bigger than the $\alpha = 0.005$, which suggests collecting more samples.

	\subsection{Comparison with the State-of-the-Art}
	
	We compare the performance of the proposed models against the state-of-the-art in lying posture detection using a single accelerometer sensor. The proposed and competing models are described as follows. 
	\begin{itemize}
		\item \textbf{ET} is the proposed feature-based classifier, which is an ensemble of decision trees trained on 48 time-domain features.
		\item \textbf{AdaLSTM} is the proposed deep learning model, which is an adaptive long short-term memory network with Adam optimizer and decaying learning rate.
		\item \textbf{LDA}, proposed by Zhang et al., is a linear discriminate analysis (LDA) classifier trained on the mean value of the signal \cite{zhang2015monitoring}.
		\item \textbf{SVM}, proposed by Jeng et al., is a multi-class linear kernel support vector machine classifier trained on the mean value of the tri-axial accelerometer signal \cite{jeng2019wrist}.
		\item \textbf{LSTM} is a long short-term memory network with the same structure as the AdaLSTM but with a fixed learning rate of 0.01.
	\end{itemize}
	
	\subsubsection{Class-Act Dataset}
	
	\tblref{CoV2} compares the F1-score mean and CoV of the proposed models AdaLSTM and ET against the state-of-the-art deep learning and feature-based models on the Class-Act dataset. The class-act dataset contains data from three major lying postures including supine, prone, and left side and nine different sensor locations. Since CoV shows the ratio of variation to mean for a metric, a lower F1-Score CoV value represents a more promising model. The linear feature-based classifiers including LDA and SVM obtain $>$88.3\% F1-Scores nd $<$0.26 CoV when applied to data from the thighs, ankles and chest, however, their performance significantly drops to 55.0\%--82.2\% F1-Score and 0.22--0.36 CoV when the sensor is moved to the arms or wrists. The competing deep learning model, LSTM, maintains 83.7\%--92.5\% mean F1-Score for the thighs, ankles and chest locations, and 51.6\%--75.5\% mean F1-Score for the arms, and wrists location. AdaLSTM outperforms the competing deep learning and feature-based classifiers over all the body locations except for the Right thigh and right wrist. It achieves 91.5\%--98.2\% mean F1-Score and 0.06--0.17 CoV for the thighs, ankles and chest body locations. It shows the most promising result when applied to the sensor on the left thigh with 98.2\% F1-Score and 0.06. These results demonstrate the power of deep learning and salient of the left thigh in detecting the lying postures for a new subject. We note that neural networks with simpler structures such as LSTM with a fixed learning rate in this paper could not extract useful features and patterns from the raw data automatically from limited data, therefore choosing the proper parameters for the deep learning models is a crucial factor in their performance.

	\begin{table}[ht]
		\centering
		\scriptsize
		\caption{Comparison between the mean value and coefficient of variation for F1-score of detecting lying postures for different sensor placements and classifiers including Ensemble Trees (ET), Linear Discriminator Analysis (LDA), LSTM with fixed learning rate (LSTM), and Adaptive LSTM (AdaLSTM). We show the highest F1-Score value, and lowest CoV metric that models could achieve for each location for LOSO validations in bold.}
		\renewcommand{\arraystretch}{1.2}
		\begin{tabular}{ccrrrrr}\cline{1-7}\cline{1-7}
			& \textbf{Location} & \textbf{ET}    & \textbf{LDA}  &\textbf{SVM} & \textbf{LSTM}   & \textbf{AdaLSTM} \\ \cline{1-7}\cline{1-7}
			\parbox[t]{4mm}{ \multirow{9}{*}{\rotatebox[origin=c]{90}{\textbf{Mean Value (\%)}}}}
			
			& \textbf{Left thigh}  &90.7& 95.4 & 95.4 &92.5	&\textbf{98.2}\\ \cline{2-7}
			& \textbf{Right thigh}	&93.5&96.1 & 93.2 &84.8	&91.5\\ \cline{2-7}
			& \textbf{Left ankle}  &92.1 &88.3 & 94.8 &90.2	&\textbf{96.9}\\ \cline{2-7}
			& \textbf{Right ankle} &82.9&90.0 & 89.5 &83.7	&\textbf{91.7} \\ \cline{2-7}
			& \textbf{Chest} &97.0	&94.8 &  90.1 &88.3 &\textbf{97.3} \\ \cline{2-7}         
			& \textbf{Left arm}   &60.9	& 58.1 & 53.3	&53.7	&\textbf{66.3}\\ \cline{2-7}
			& \textbf{Right arm}   &81.6& \textbf{82.2} &  76.1	&75.5	&75.7 \\ \cline{2-7}
			& \textbf{Left wrist}  &\textbf{64.1} &55.0 &  50.7	&51.6	& 64.0\\ \cline{2-7}
			& \textbf{Right wrist} 	&67.9& 65.5 &  59.2	&54.1	&\textbf{69.4} \\ \cline{1-7}
			
			\parbox[t]{8mm}{\multirow{9}{*}{\rotatebox[origin=c]{90}{\begin{tabular}[c]{@{}l@{}}\textbf{Coefficient of}\\  \textbf{variation}\end{tabular}}}} 
			
			& \textbf{Left thigh}  	& 0.13	&0.15 & 0.13  &0.22	&\textbf{0.06}\\ \cline{2-7}
			& \textbf{Right thigh} 	& \textbf{0.15}	&0.16 & 0.17   &0.28 &0.17\\ \cline{2-7}
			& \textbf{Left ankle}  	& 0.17	&0.18 & 0.16   &0.24 &\textbf{0.06}\\ \cline{2-7}
			& \textbf{Right ankle} 	& 0.27	&0.18 &  0.12   &0.27 &\textbf{0.11}\\\cline{2-7}
			& \textbf{Chest} & 0.17	&0.16   &0.26 & 0.15    &\textbf{0.07}\\\cline{2-7}
			& \textbf{Left arm}  & 0.27	&0.33 &  0.29 & 0.36  & \textbf{0.24}\\\cline{2-7}
			& \textbf{Right arm}& 0.26	&\textbf{0.22} & 0.24 &0.23	&\textbf{0.22}\\\cline{2-7}
			& \textbf{Left wrist}& 0.33	&0.36 & 0.36 &0.50	&\textbf{0.25}\\\cline{2-7}
			& \textbf{Right wrist}	& 0.35	&0.32 &  \textbf{0.29}	&0.36	&0.31\\ \cline{1-7}
			
		\end{tabular}
		\label{tbl:CoV2}
	\end{table}

	\figref{conf_et}, and \figref{conf_lstm} show the confusion matrix of the ET and AdaLSTM classifiers for the sensor on the thighs, and the wrists locations. As shown, both classifiers mislabel a few of the episodes when applied to data from the sensor on the thighs.

	\subsubsection{Integrated Dataset}
	\tblref{CoV4} shows the mean F1-score and CoV values of lying posture detection on the dataset of four major lying postures including supine, prone, left side and right side and five sensor locations including thighs, wrists, and chest.  The results are leave-one-subject-out validated because it is a more realistic validation scenario for the application of human lying posture tracking. 
	
	As shown in \tblref{CoV4}, ET and AdaLSTM classifiers achieve a promising range of F1-Score (i.e., $63.3\%$ for the right wrist to $97.0\%$ for the chest) across all the body locations. ET classifier obtains the highest mean F1-score when the sensor is placed on the right thigh (i.e., $97.3\%$) and right wrist (i.e., $78.6\%$) locations, while AdaLSTM achieves the highest F1-Score values for the chest (i.e., $95.3\%$) and the right wrist(i.e., $74.2\%$) among all the algorithms. The linear classifiers such as LDA \cite{zhang2015monitoring} and SVM \cite{jeng2019wrist} achieve higher F1-Score (i.e ranged 90.4\%-97.2\%) than the proposed models when applied to data collected from the sensor on the thighs and the chest. The linear relationship between the lying posture and accelerometer readings causes the superiority of state-of-the-art for these locations. On the other hand, extra movements of the hands during lying introduce noise and non-linearity to the data collected by the sensor placed on these locations, therefore, F1-Score values of linear classifiers drop to $33.5\%$, and $44.6\%$ for the left and right wrists. 
	
	Moreover, the proposed models show lower F1-Score variation to mean ratio comparing to the state-of-the-art techniques. Ensemble tree classifier achieves CoV of $0.14$, and $0.15$, for the right thigh, the chest, respectively, and AdaLSTM achieves CoV of $0.39$, and $0.44$ for the left and the right wrist, respectively. While CoV of the linear models including LDA and SVM increases to the range $0.53-0.80$ for the left and right wrist locations.
	
	\begin{table}[ht]
		\centering
		\scriptsize
		\caption{Comparison between the mean value and coefficient of variation for F1-score of detecting lying postures for different sensor placements and classifiers including Ensemble Trees (ET), Linear Discriminator Analysis (LDA), Support Vector Machine (SVM), LSTM with fixed learning rate (LSTM), and Adaptive LSTM (Ada-LSTM) for leave-one-subject-out validation.}
		\renewcommand{\arraystretch}{1.2}
		\begin{tabular}{ccrrrrr}\cline{1-7}\cline{1-7}
    		& \textbf{Location} & \textbf{ET} & \textbf{LDA} & \textbf{SVM}   & \textbf{LSTM} & \textbf{AdaLSTM}\\ \cline{1-7}\cline{1-7}
    		
			\parbox[t]{10mm}{ \multirow{5}{*}{\rotatebox[origin=c]{90}{\begin{tabular}[c]{@{}l@{}}\textbf{Mean}\\  \textbf{Value($\%$)}\end{tabular}}}}

			& \textbf{Left thigh} & 90.6 & \textbf{94.6} & 91.4 & 92.9 & 93.7\\ \cline{2-7}
			& \textbf{Right thigh} 	& \textbf{97.3} & 96.9 & 91.4 & 93.2 & 94.0\\ \cline{2-7}
			& \textbf{Chest} & 95.4	& 95.4  & \textbf{96.7}	&90.7 & 95.0\\ \cline{2-7}         
			& \textbf{Left wrist} & \textbf{65.9} & 42.1 & 58.4 & 54.1 & 63.3\\ \cline{2-7}
			& \textbf{Right wrist} & \textbf{78.6} & 66.7 & 74.3  & 42.1 & 69.2 \\ \cline{1-7}
			\parbox[t]{8mm}{\multirow{5}{*}{\rotatebox[origin=c]{90}{\begin{tabular}[c]{@{}l@{}}\textbf{Coefficient}\\  \textbf{of variation}\end{tabular}}}} 

			& \textbf{Left thigh}  & 0.19	&0.14 &0.25 &0.09	&0.21\\ \cline{2-7}
			& \textbf{Right thigh} & \textbf{0.13}&0.15&0.20 &0.16	&0.17\\ \cline{2-7}
			& \textbf{Chest}& 0.13	&0.13	&0.24 &0.25	& \textbf{0.12}\\\cline{2-7}
			& \textbf{Left wrist}  & \textbf{0.39}	& 0.50	& 0.50 & 0.53 & 0.42\\\cline{2-7}
			& \textbf{Right wrist} & 0.34 & 0.38 & 0.32 & 0.40 & 0.39\\ \cline{1-7}
			
		\end{tabular}
		\label{tbl:CoV4}
	\end{table}

	\section{Discussion}
\begin{figure*}[ht]
	\centering
	\begin{subfigure}{.33\linewidth}
	    \includegraphics[width=\linewidth]{figures/cbar.pdf}
	\end{subfigure}
	
		\begin{subfigure}{.45\linewidth}
			\includegraphics[width=\linewidth]{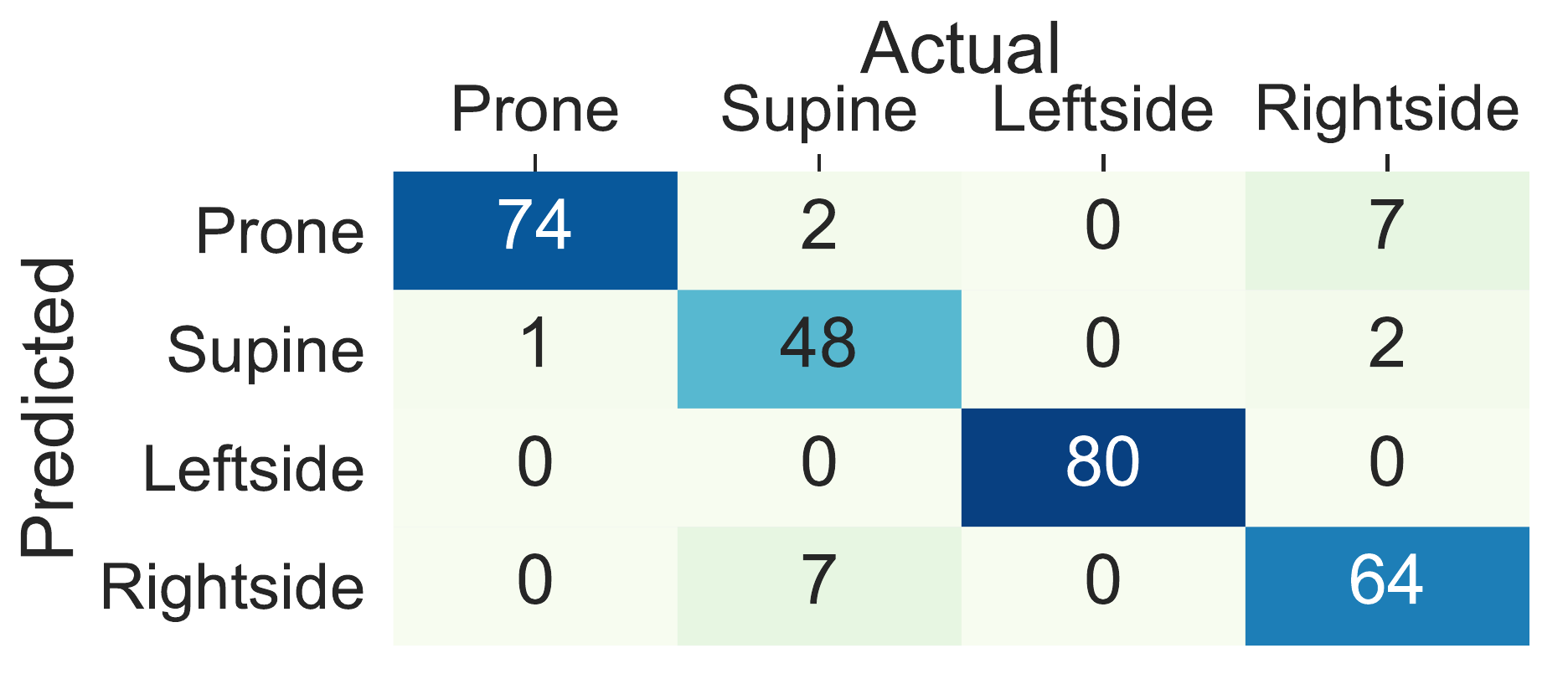}
			\caption{Left thigh}
			\label{fig:lstm_lt_int}
		\end{subfigure}
		\begin{subfigure}{.45\textwidth}
			\includegraphics[width=\linewidth]{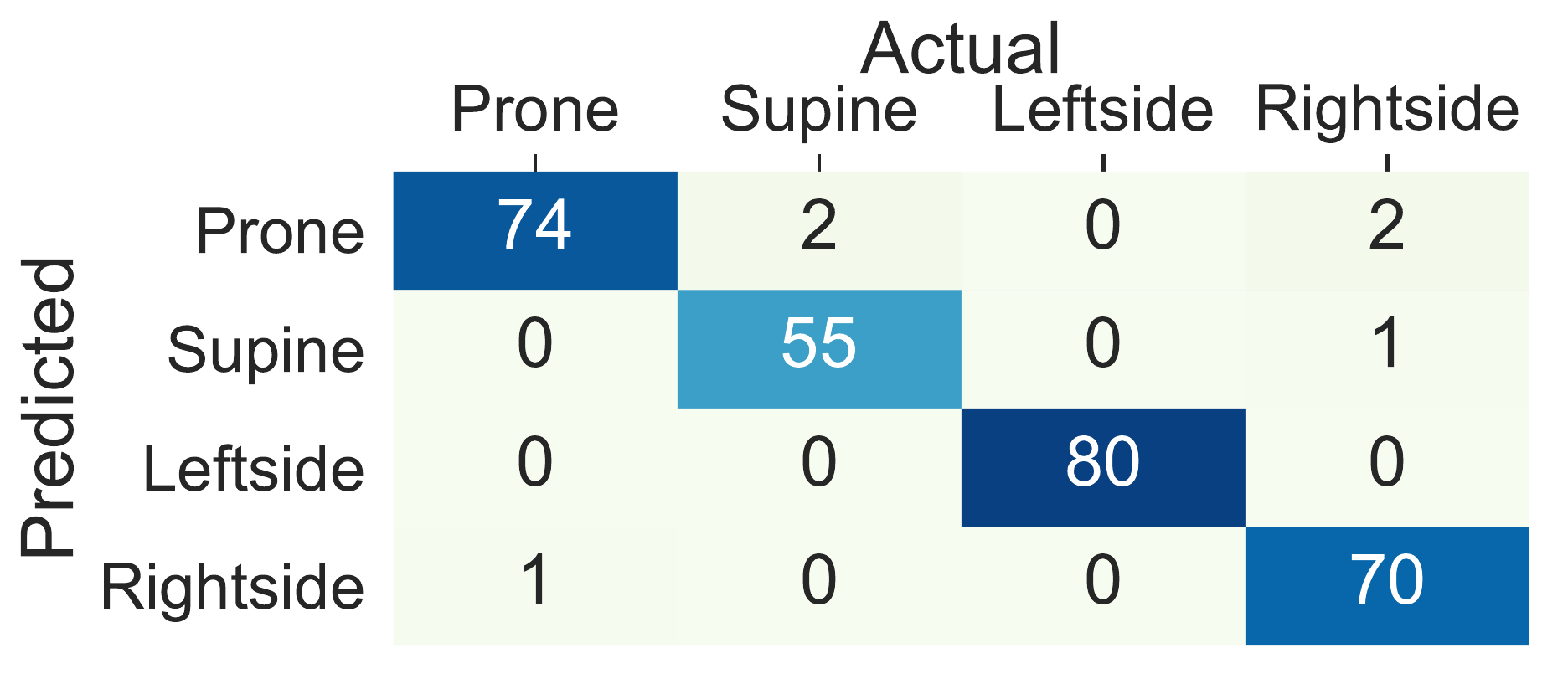}
			\caption{Right thigh}
			\label{fig:lstm_rt_int}
		\end{subfigure}
		\begin{subfigure}{.45\textwidth}
			\includegraphics[width=\linewidth]{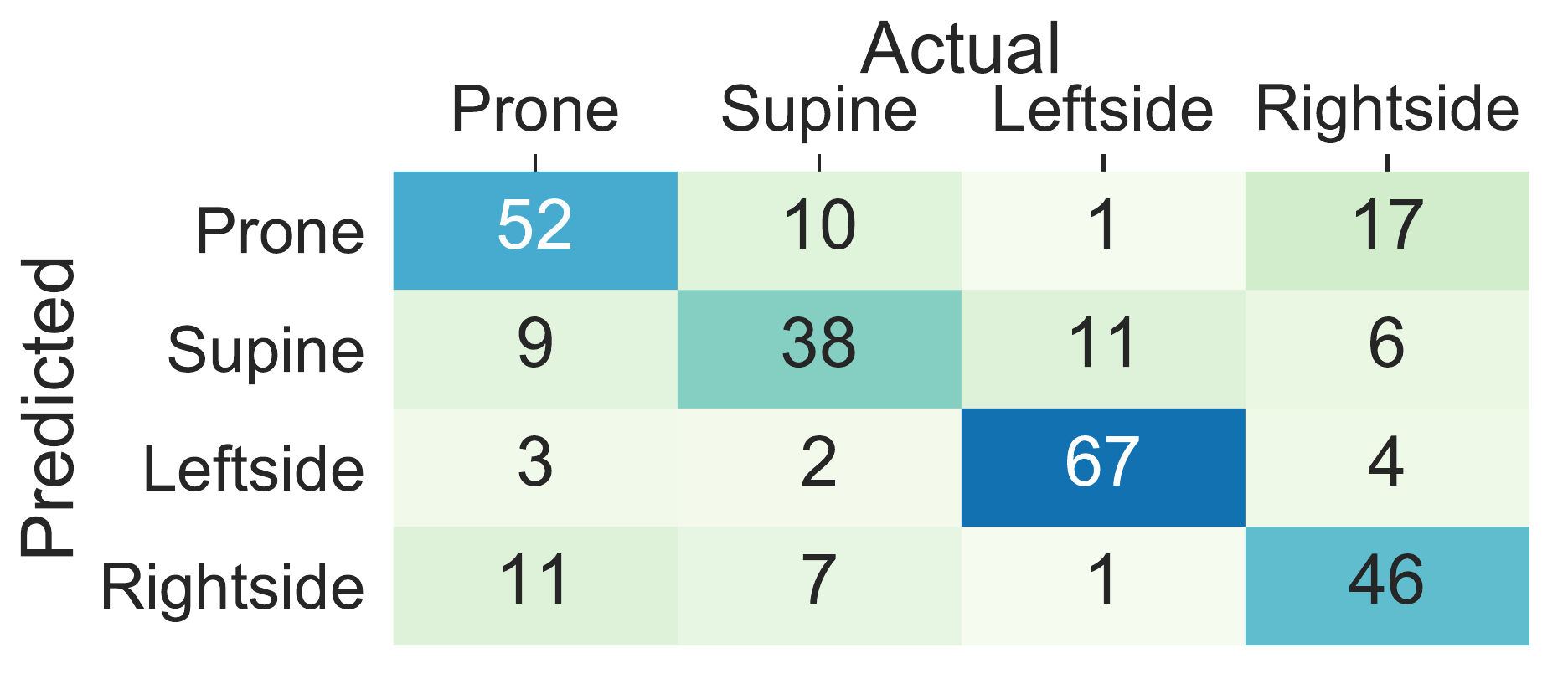}
			\caption{Left Wrist}
			\label{fig:lstm_lw_int}
		\end{subfigure}
		\begin{subfigure}{.45\textwidth}
			\includegraphics[width=\linewidth]{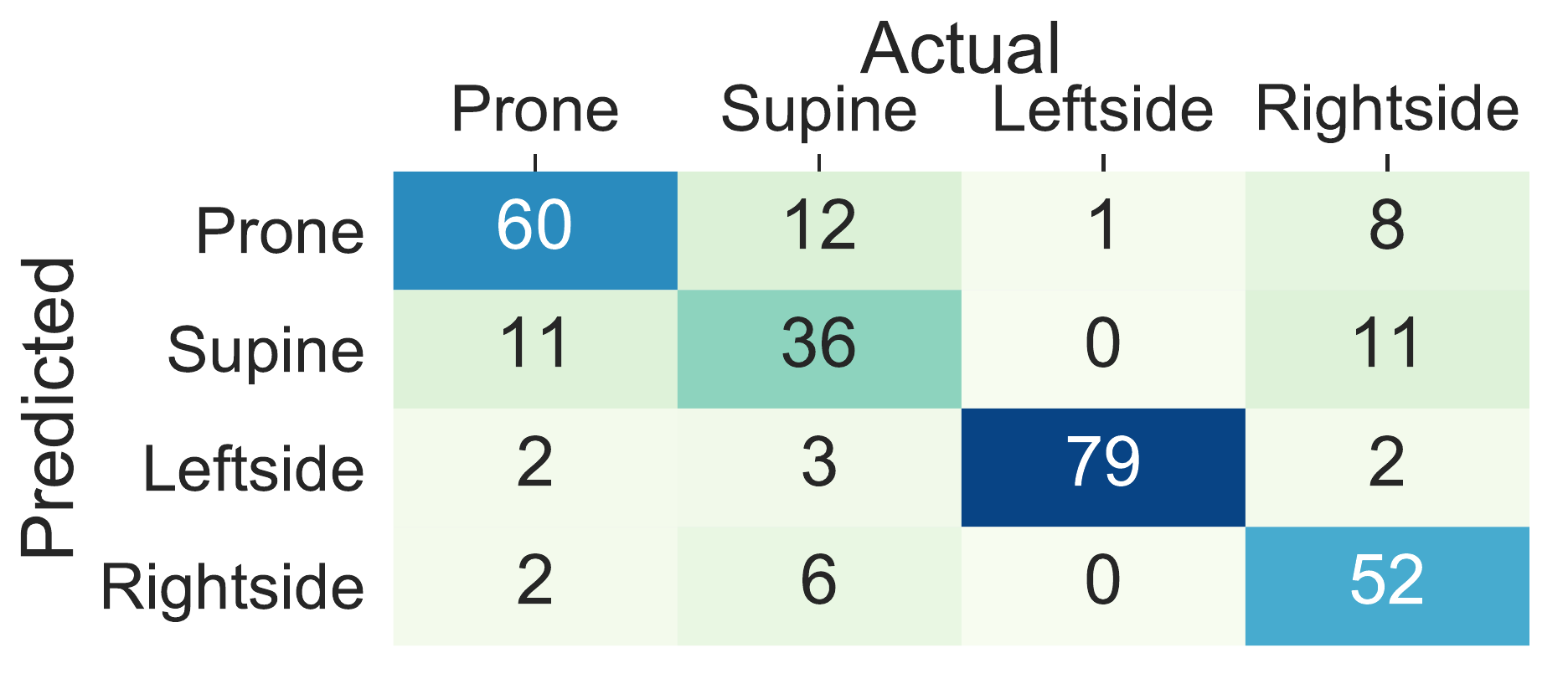}
			\caption{Right wrist}
			\label{fig:lstm_rw_int}
		\end{subfigure}
		\begin{subfigure}{.45\textwidth}
			\includegraphics[width=\linewidth]{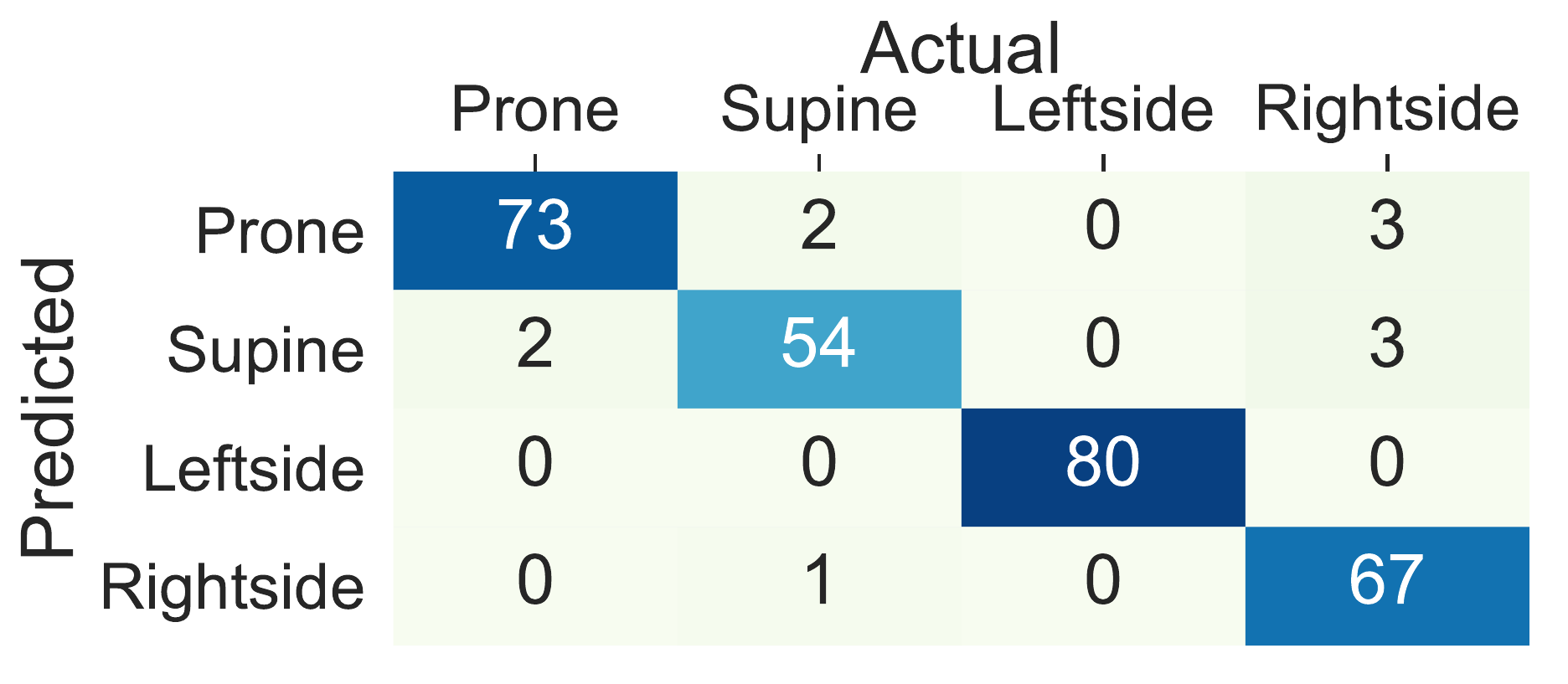}
			\caption{Chest}
			\label{fig:lstm_c_int}
		\end{subfigure}
		\caption{Confusion matrix of ensemble tree classifier in classifying lying postures into supine, prone, and left side for the thighs, ankles, arms, and wrists locations.}
		\label{fig:conf_et_int}
	\end{figure*}

	We compared the accuracy of lying posture tracking of nine different body locations in this study. When the ensemble tree classifier is trained on the data collected from the sensors on the chest and thighs the lying posture tracking achieves the highest performance and the least cross-subject variations, while the wrists and the arms classifiers show the least performance and highest within-subject variations. These results demonstrate that individuals might devise arbitrary and dissimilar hand movements during the same lying posture. \figref{conf_et_int} Compares the confusion matrices of the ensemble tree classifiers trained on the data from the chest, thighs, and wrists from integrated dataset. The chest left thigh, and the right thigh classifiers confuse 3.7\%, 6.6\% and 2.1\% of the lying episodes, respectively, While this ratio increases to 55.1\% for the left wrist and 61.4\% for the right wrist sensor. The confusion between the supine and prone postures in the wrists' sensors is caused by the wrist rotations while lying. In particular, The right side posture is mainly confused with the prone when the sensor is on the left wrist and confused with the supine when the sensor is on the right wrist. Moreover, the majority of the confusions between the right side and supine postures occur when the sensor is on the right wrist, and the left side and prone posture occur when the sensor in on the left wrist. These results are mainly due to the similar sensor position during the postures that are confused with each other. For example, the left wrist holds similar positions when the user lays on the back (i.e., supine) and lays on the left side depending on the rotation of the wrist. Also, the right wrist might adopt the same position when the user lays on the front (prone) and lays on the right side.
	
\begin{figure*}[ht]
	\centering
			\begin{subfigure}{.33\linewidth}
	    \includegraphics[width=\linewidth]{figures/cbar.pdf}
	\end{subfigure}
	
		\begin{subfigure}{.45\linewidth}
			\includegraphics[width=\linewidth]{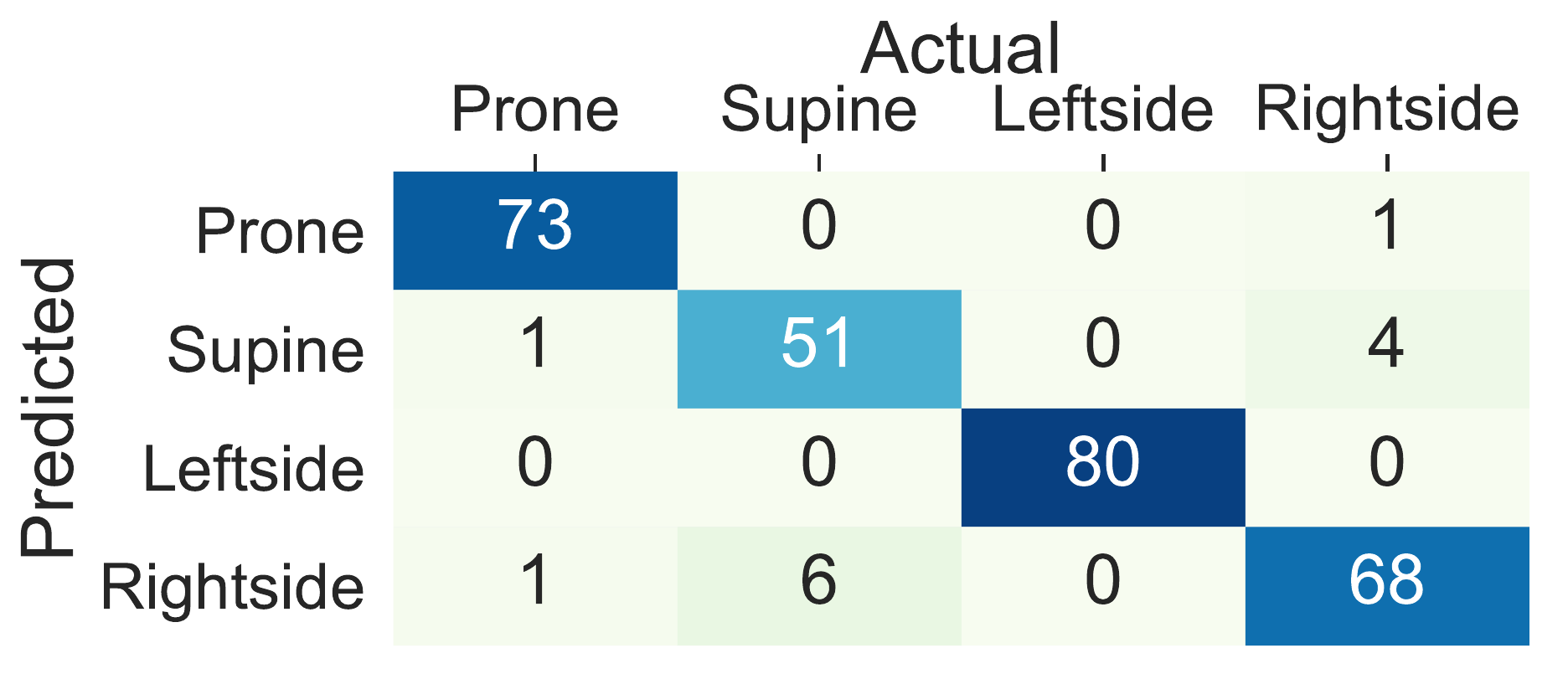}
			\caption{Left thigh}
			\label{fig:lstm_lt_int}
		\end{subfigure}
		\begin{subfigure}{.45\textwidth}
			\includegraphics[width=\linewidth]{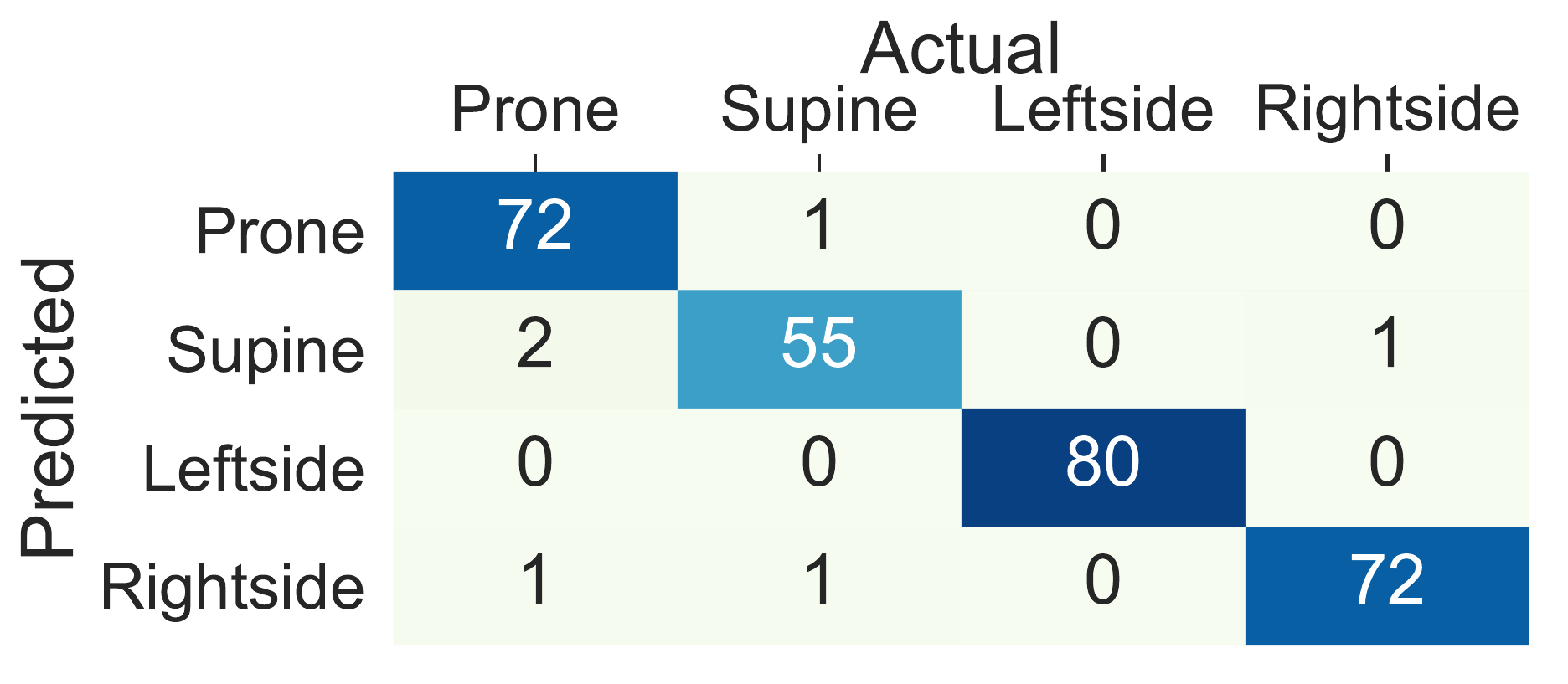}
			\caption{Right thigh}
			\label{fig:lstm_rt_int}
		\end{subfigure}
		\begin{subfigure}{.45\textwidth}
			\includegraphics[width=\linewidth]{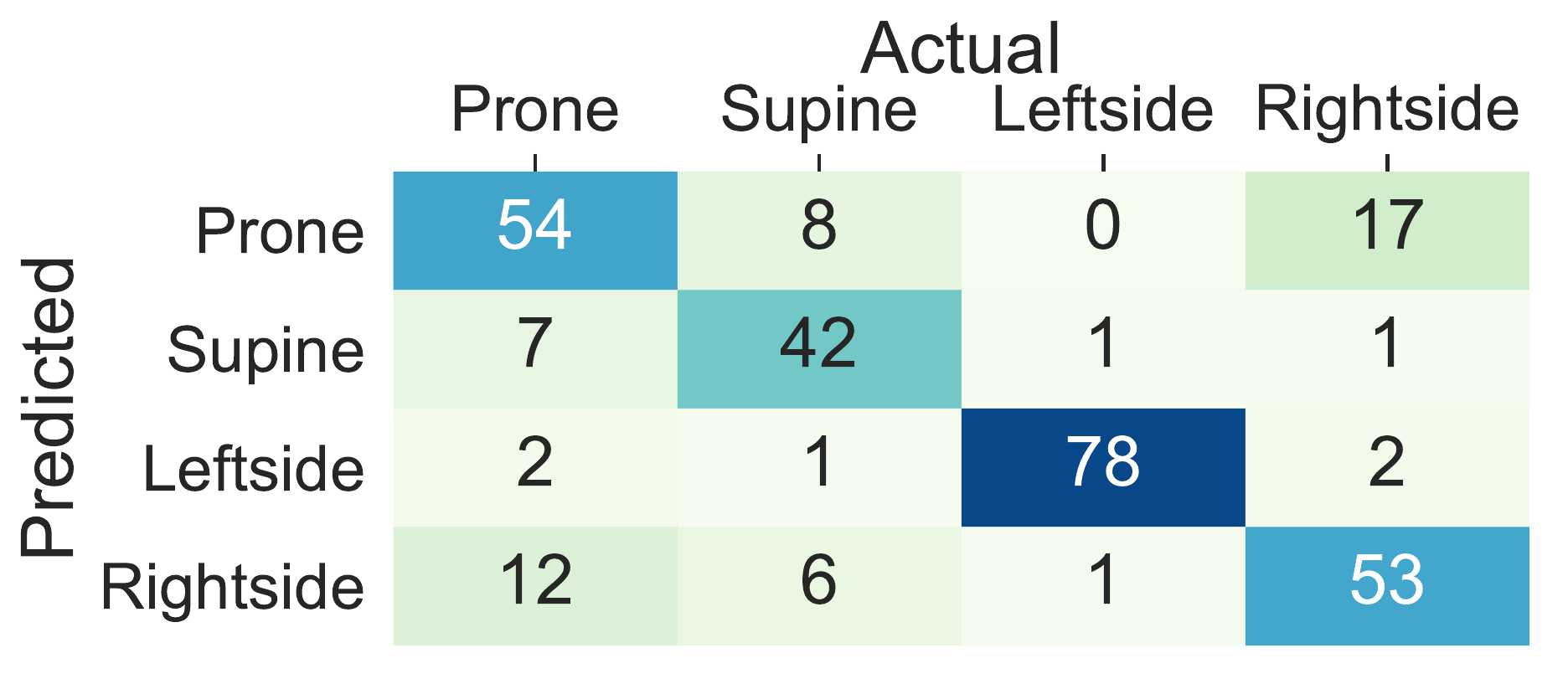}
			\caption{Left Wrist}
			\label{fig:lstm_lw_int}
		\end{subfigure}
		\begin{subfigure}{.45\textwidth}
			\includegraphics[width=\linewidth]{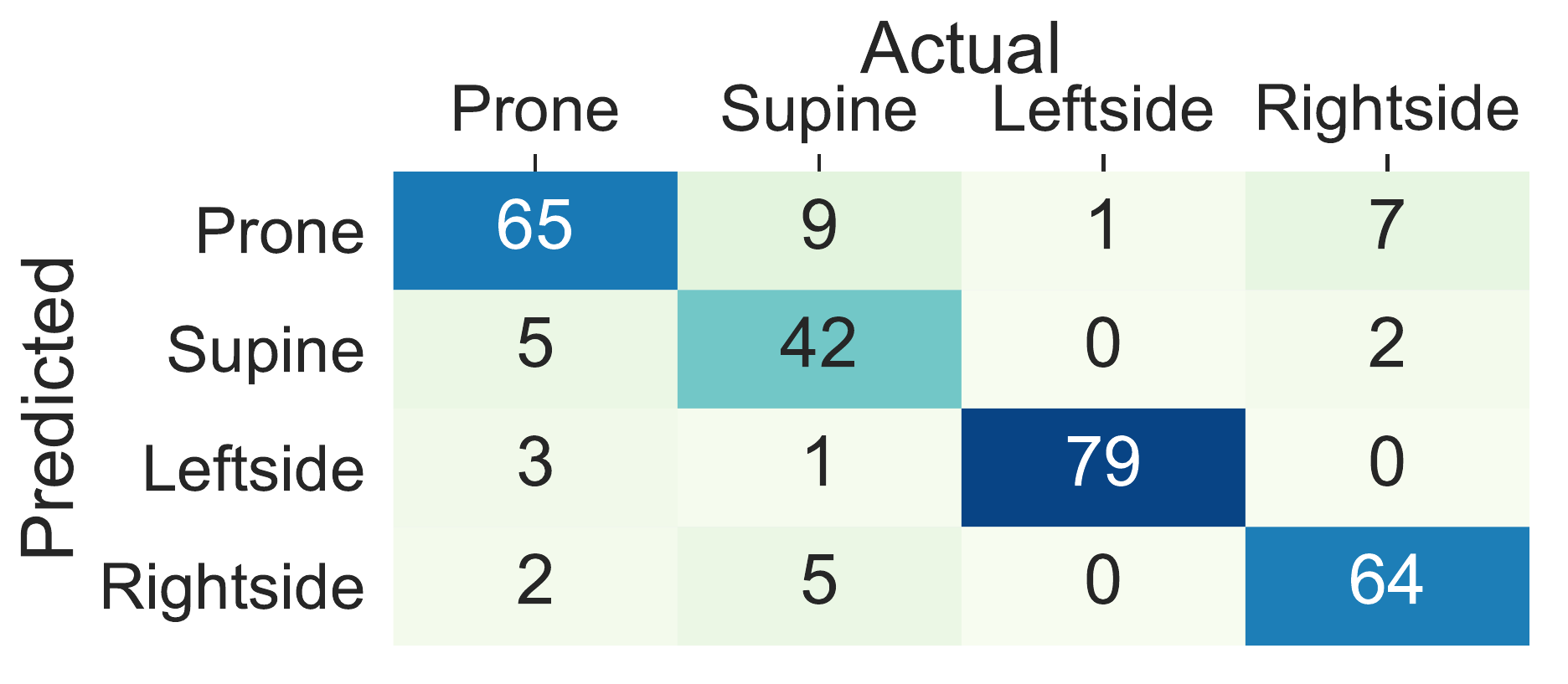}
			\caption{Right wrist}
			\label{fig:lstm_rw_int}
		\end{subfigure}
		\begin{subfigure}{.45\textwidth}
			\includegraphics[width=\linewidth]{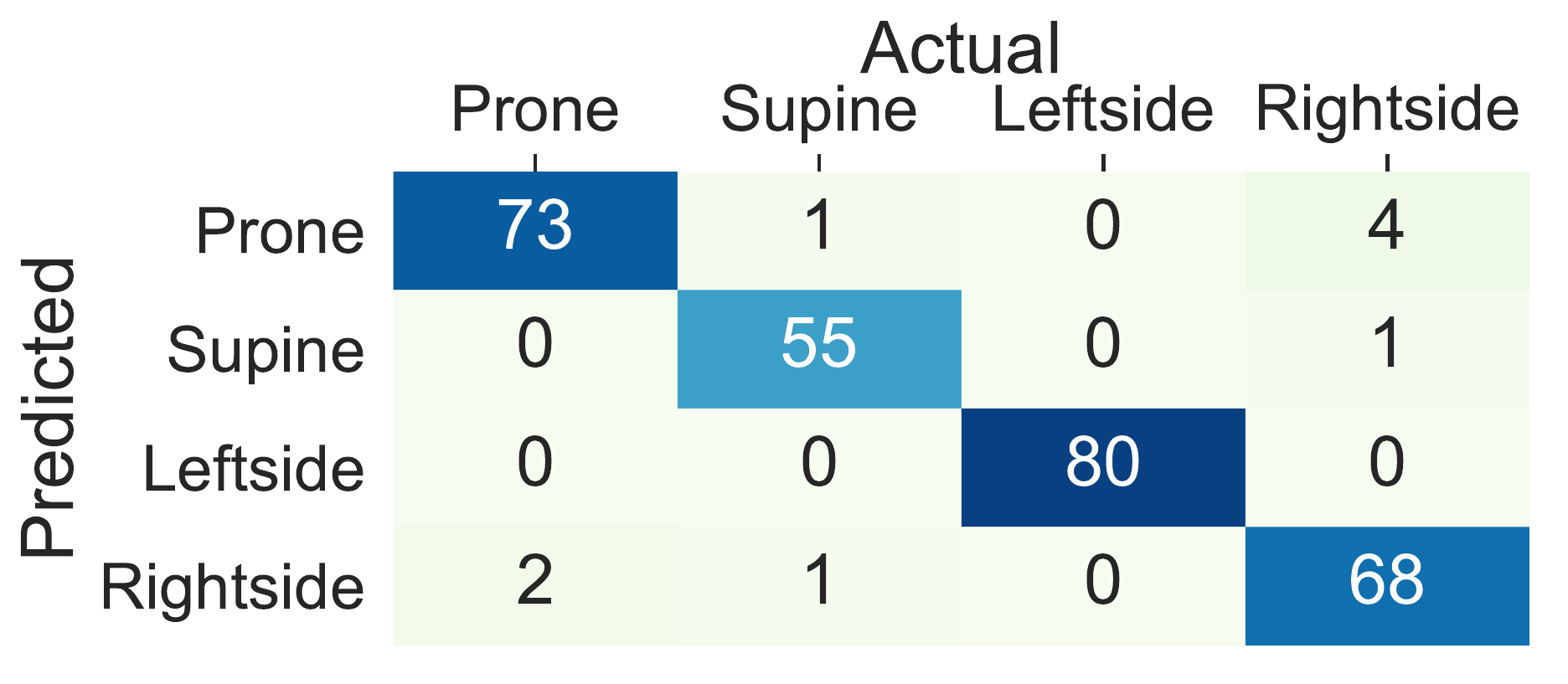}
			\caption{Chest}
			\label{fig:lstm_c_int}
		\end{subfigure}

		\caption{Confusion matrix of AdaLSTM classifier in classifying lying postures into supine, prone, and left side for the thighs, ankles, arms, and wrists locations.}
		\label{fig:conf_lstm_int}
	\end{figure*}

	We further investigated the possibility of replacing traditional machine learning with deep learning. Our study showed that deep RNNs such as LSTM can replace the traditional machine learning classifiers as long as adequately designed. \figref{conf_lstm_int} shows the confusion matrices for lying posture detection using AdaLSTM on data collected from the sensor on the chest, the thighs, and the wrists. These results follow a similar trend as the ensemble tree classifier. 3.1\%, 4.5\%, and 2.1\% of the lying episodes are misclassified when the sensor is worn on the chest, the left thigh, and the right thigh, respectively. While the misclassification rate increases to 20.3\% and 12.2\% for the left wrist, and the right wrist classifiers respectively. The AdaLSTM achieve confuses 34.8\%, and 49.2\% less lying episodes compared to the ensemble tree classifier when the sensor is placed on the left wrist, and the right wrist, respectively. These results show the ability of the deep RRNS to capture non-linear relations in the data based on the nonlinear operations on a higher level of abstraction. In addition, deep RRNS  such as AdaLSTM do not require feature-engineering. One major drawback of deep learning is the inability to interpret extracted features through the deeper layers of the network. Moreover, these models are computationally expensive and require large datasets for training to achieve promising results \cite{gamboa2017deep}. 

	The fact that end-to-end deep learning neural networks could not improve the performance significantly comparing to the feature-based classifiers demonstrates the lack of sufficient data as a limitation to this study \cite{goodfellow2018atrial}. We believe adding more data to the training dataset will further improve the performance of AdaLSTM especially for the data from the sensor on the wrists and the arms of the users. We are planning to address this issue in two directions: (1) conduct an extensive multi-modality data collection from a large number of participants performing different lying postures, including main postures and their other variations; (2) produce signal-/feature-level synthesis data using data augmentation techniques such as rotation, permutation, time-wrapping, scaling, magnitude-wrapping, jittering \cite{fawaz2018data}, sequence to sequence learning techniques \cite{devries2017dataset}, and generative adversarial networks \cite{wang2018sensorygans}.
		
\section{Conclusion}
	We implemented a traditional machine learning classifier, ensemble tree with time-domain features, and a deep recurrent neural network, AdaLSTM, with decaying learning rate to detect four major lying postures including, supine, prone, left side and right side, using a single tri-axial accelerometer sensor. We identified amplitude, mean, minimum, and maximum values of the lateral and vertical axes as the optimal set of time-domain features for accurate lying posture tracking using a single accelerometer sensor. We determined the optimal wearing sites of a single accelerometer sensor to accurately detect lying postures. Finally, we evaluated the performance of the proposed models against deep learning and feature-based state-of-the-art lying posture tracking methods using two publicly available human posture and activity datasets. The proposed AdaLSTM using data from the left thigh and AdaLSTM on the chest locations achieved the highest F1-Scores ($98.5\%$ for the left thigh and $97.8\%$ for the chest) and lowest coefficient of variations ($0.07$ for the left thigh and $0.03$ for the chest) compared to the other models and sensor locations for the Class-Act dataset. The proposed ensemble tree classifier achieved $97.1\%$ F1-Score and $0.14$ CoV when applied to the data from the sensor on the right thigh, and AdaLSTM obtained $95.3\%$ F1-Score and $0.15$ CoV when applied to the data from the chest sensor from Integration of Class-Act and DAS datasets including 20 subjects. These results demonstrated the thighs and the chest as the optimum location for the accelerometer sensor for accurate lying posture tracking. 
	
\bibliographystyle{elsarticle-num-names}
\bibliography{sample}
\end{document}